\documentclass[sigconf]{acmart}

\usepackage{booktabs} 
\usepackage{algorithm}
\usepackage[noend]{algpseudocode}
\usepackage{amsmath,amsthm,amssymb}
\usepackage{graphicx}
\usepackage{enumitem}
\usepackage{setspace}
\usepackage{color}
\usepackage[normalem]{ulem} 
\usepackage{subfigure} 
\theoremstyle{definition}

\usepackage{multirow}
\usepackage{subfigure} 

\usepackage{ textcomp }
\usepackage{ gensymb }
\usepackage{ wasysym }
\setcopyright{rightsretained}

\acmDOI{nn.nnn/nnn_n}

\acmISBN{nnn-nnnn-nn-nnn/nn/nn}

\acmConference[SIGIR'19]{ACM conference}{July 2019}{}
\acmYear{2019}
\copyrightyear{2019}

\acmArticle{4}
\acmPrice{15.00}

 \newcommand{\squishlist}{
	\begin{list}{$\bullet$}
		{ \setlength{\itemsep}{0pt}
			\setlength{\parsep}{3pt}
			\setlength{\topsep}{3pt}
			\setlength{\partopsep}{0pt}
			\setlength{\leftmargin}{1.5em}
			\setlength{\labelwidth}{1em}
			\setlength{\labelsep}{0.5em} } }
	
	\newcommand{\squishlisttwo}{
		\begin{list}{$\bullet$}
			{ \setlength{\itemsep}{0pt}
				\setlength{\parsep}{0pt}
				\setlength{\topsep}{0pt}
				\setlength{\partopsep}{0pt}
				\setlength{\leftmargin}{2em}
				\setlength{\labelwidth}{1.5em}
				\setlength{\labelsep}{0.5em} } }
		
		\newcommand{\squishend}{
	\end{list}  }

	\newfont{\mycrnotice}{ptmr8t at 7pt}
	\newfont{\myconfname}{ptmri8t at 7pt}

\begin{document}
	\copyrightyear{2019}
	\acmYear{2019}
	\setcopyright{acmcopyright}
	\acmConference[SIGIR '19]{Proceedings of the 42nd International ACM SIGIR Conference on Research and Development in Information Retrieval}{July 21--25, 2019}{Paris, France}
	\acmBooktitle{Proceedings of the 42nd International ACM SIGIR Conference on Research and Development in Information Retrieval (SIGIR '19), July 21--25, 2019, Paris, France}
	\acmPrice{15.00}
	\acmDOI{10.1145/3331184.3331248}
	\acmISBN{978-1-4503-6172-9/19/07}
\title[Learning from Fact-checkers: Analysis and Generation of Fact-checking Language ]{Learning from Fact-checkers: Analysis and Generation \\ of Fact-checking Language}




%
\author{Nguyen Vo}
\affiliation{%
  \institution{Worcester Polytechnic Institute}
  \city{Worcester}
  \state{Massachusetts}
  \country{USA, 01609}
}
\email{nkvo@wpi.edu}

\author{Kyumin Lee}
\affiliation{%
 \institution{Worcester Polytechnic Institute}
 \city{Worcester}
 \state{Massachusetts}
 \country{USA, 01609}}

\email{kmlee@wpi.edu}
%
%
%


\begin{abstract}
	
	In fighting against fake news, many fact-checking systems comprised of human-based fact-checking sites (e.g., snopes.com and politifact.com) and automatic detection systems have been developed in recent years. However, online users still keep sharing fake news even when it has been debunked. It means that early fake news detection may be insufficient and we need another complementary approach to mitigate the spread of misinformation. In this paper, we introduce a novel application of text generation for combating fake news. In particular, we (1) leverage online users named \emph{fact-checkers}, who cite fact-checking sites as credible evidences to fact-check information in public discourse; (2) analyze linguistic characteristics of fact-checking tweets; and (3) propose and build a deep learning framework to generate responses with fact-checking intention to increase the fact-checkers' engagement in fact-checking activities. Our analysis reveals that the fact-checkers tend to refute misinformation and use formal language (e.g. few swear words and Internet slangs). Our framework successfully generates relevant responses, and outperforms competing models by achieving up to 30\% improvements. Our qualitative study also confirms that the superiority of our generated responses compared with responses generated from the existing models.







\end{abstract}

%
%



\maketitle

\section{Introduction}


Our media landscape has been flooded by a large volume of falsified information, overstated statements, false claims, fauxtography and fake videos\footnote{\url{https://cnnmon.ie/2AWCCix}} perhaps due to the popularity, impact and rapid information dissemination of online social networks. The unprecedented amount of disinformation posed severe threats to our society, degraded trustworthiness of cyberspace, and influenced the physical world. For example, \$139 billion was wiped out when the Associated Press (AP)'s hacked twitter account posted fake news regarding White House explosion with Barack Obama's injury.

To fight against fake news, many fact-checking systems ranging from human-based systems (e.g. Snopes.com), classical machine learning frameworks \cite{kwon2013aspects,popat2016credibility,nguyen2018interpretable} to deep learning models \cite{ma2016detecting,wang2017liar,wang2018eann,popat2018declare} were developed to determine credibility of online news and information. However, falsified news is still disseminated like wild fire \cite{maddock2015characterizing, zhao2015enquiring} despite dramatic rise of fact-checking sites worldwide \cite{reporterLabFCURLs}. 
 Furthermore, recent work showed that individuals tend to selectively consume news that have ideologies similar to what they believe while disregarding contradicting arguments \cite{ecker2010explicit, nyhan2010corrections}. These reasons and problems indicate that using only fact-checking systems to debunk fake news is insufficient, and complementary approaches are necessary to combat fake news.

\begin{figure}[t]
	\centering
	\includegraphics[trim=50 210 350 70,clip,width=\linewidth,height=1.7in]{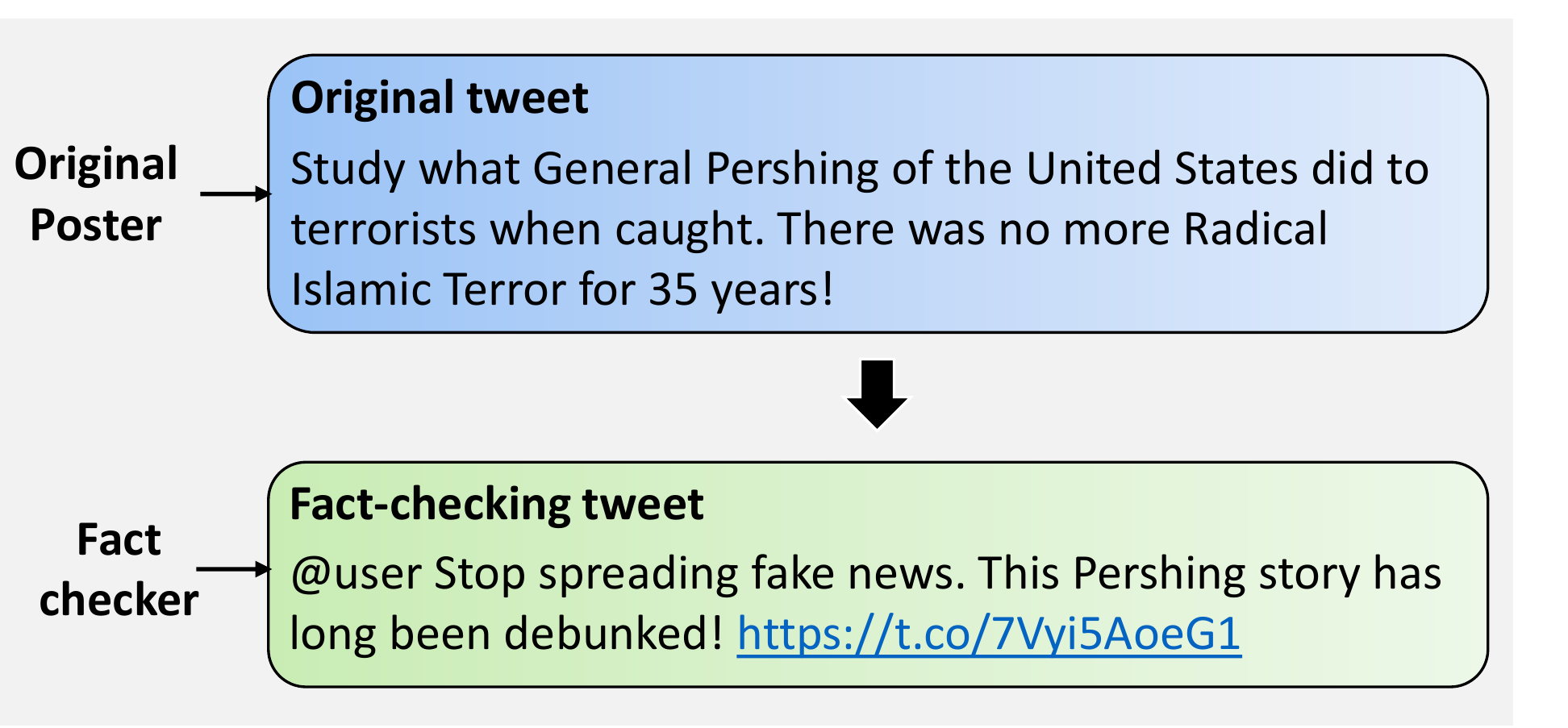}
	\caption{A real-life fact-checking activity where the fact-checker refutes misinformation in the original tweet. }
	\label{fig:example2}
	\vspace{-10pt}
\end{figure}

Therefore, in this paper, we focus on online users named \textit{fact-checkers}, who directly engage with other users in public dialogues and convey verified information to them. Figure \ref{fig:example2} shows a real-life conversation between a user, named \textit{original poster}, and a fact-checker. In Figure \ref{fig:example2}, the original poster posts a false claim related to General Pershing. The \textit{fact-checker} refutes the misinformation by replying to the original poster and provides a fact-checking article as a supporting evidence. We call such a reply a \textit{fact-checking tweet} (FC-tweet). Recent work \cite{vo2018rise} showed that fact-checkers often quickly fact-checked original tweets within a day after being posted and their FC-tweets could reach hundreds of millions of followers. Additionally, \cite{friggeri2014rumor} showed that the likelihood to delete shares of fake news increased by four times when there existed a fact-checking URL in users' comments. In our analysis, we also observe that after receiving FC-tweets, 7\% original tweets were not accessible because of account suspension, tweet deletion, and a private mode.


Due to the fact-checkers' activeness and high impact on dissemination of fact-checked content, in this paper, our goal is to further support them in fact-checking activities toward complementing existing fact-checking systems and combating fake news. In particular, we aim to build a text generation framework to generate responses\footnote{We use the term ``fact-checking tweets (FC-tweets)'', ``fact-checking responses'', and ``fact-checking replies'' interchangeably.} with fact-checking intention when original tweets are given. The fact-checking intention means either confirming or refuting content of an original tweet by providing credible evidences. We assume that fact-checkers choose the fact-checking URLs by themselves based on their interests (e.g., \url{https://t.co/7Vyi5AoeG1} in Figure \ref{fig:example2}). Therefore, we focus on generating responses without automatically choosing specific fact-checking URLs, which is beyond the scope of this paper. 

To achieve the goal, we have to solve the following research problems: (\textbf{P1}) how can we obtain a dataset consisting of original tweets and associated fact-checking replies (i.e., replies which exhibit fact-checking intention)?; (\textbf{P2}) how can we analyze how fact-checkers communicate fact-checking content to original posters?; and (\textbf{P3}) how can we automatically generate fact-checking responses when given content of original tweets?

To tackle the first problem \textbf{(P1)}, 
we may use already available datasets \cite{jiang2018linguistic,vosoughi2018spread,vo2018rise}. However, the dataset in \cite{jiang2018linguistic} contains relatively small number of original tweets ($\sim$5,000) and many FC-tweets ($\sim$170K). Since FC-tweet generation process depends on contents of original tweets, it may reduce diversity of generated responses. The dataset in \cite{vosoughi2018spread} is large but fully anonymized, and the dataset in \cite{vo2018rise} does not contain original tweets. Therefore, we collected our own dataset consisting of 64,110 original tweets and 73,203 FC-tweets (i.e., each original tweet receives 1.14 FC-tweet) by using Hoaxy system \cite{shao2016hoaxy} and FC-tweets in \cite{vo2018rise}. 


To understand how fact-checkers convey credible information to original posters and other users in online discussions (\textbf{P2}), we conducted data-driven analysis of FC-tweets and found that fact-checkers tend to refute misinformation and employ more impersonal pronouns. Their FC-tweets were generally more formal and did not contain much swear words and Internet slangs. These analytical results are important since we can reduce the likelihood to generate racist tweets \cite{MicrosoftBot}, hate speeches \cite{davidson2017automated} and trolls \cite{cheng2017anyone}. 

To address the third problem (\textbf{P3}), we propose a deep learning framework to automatically generate fact-checking responses for fact-checkers. In particular, we build the framework based on Seq2Seq \cite{sutskever2014sequence} with attention mechanisms \cite{luong2015effective}.

Our contributions are as follows:
\squishlist
\item 
To the best of our knowledge, we are the first to propose a novel application of text generation for supporting fact-checkers and increasing their engagement in fact-checking activities. 
\item We conduct a data-driven analysis of linguistic dimensions, lexical usage and semantic frames of fact-checking tweets. 
\item We propose and build a deep learning framework to generate responses with fact-checking intention. Experimental results show that our models outperformed competing baselines quantitatively and qualitatively.
\item We release our collected dataset and source code in public to stimulate further research in fake news intervention\footnote{\url{https://github.com/nguyenvo09/LearningFromFactCheckers}}. 
\squishend

\section{Related work}
In this section, we briefly cover related works about (1) misinformation and fact-checking, and (2) applications of text generation.

\subsection{Misinformation and Fact-checking}
Fake news is recently emerging as major threats of credibility of information in cyberspace. Since human-based fact-checking sites could not fact-check every falsified news, many automated fact-checking systems were developed to detect fake news in its early stage by using different feature sets \cite{qazvinian2011rumor,gupta2013faking,zhao2015enquiring,vosoughi2018spread,jiang2018linguistic}, knowledge graph \cite{shiralkar2017finding} and crowd signals \cite{nguyen2018interpretable,kim2018leveraging,kim2019homogeneity}, and using deep learning models \cite{ma2015detect,wang2018eann,popat2018declare}. In addition, other researchers studied how to fact-check political statements \cite{wang2017liar}, mutated claims from Wikipedia \cite{thorne2018fever} and answers in Q\&A sites \cite{mihaylova2018fact}.

Other researchers studied intention of spreading fake news (e.g. misleading readers, inciting clicks for revenue and manipulating public opinions) and different types of misinformation (e.g. hoaxies, clickbait, satire and disinformation) \cite{volkova2017separating,rashkin2017truth}. Linguistic patterns of political fact-checking webpages and fake news articles \cite{rashkin2017truth,horne2017just} were also analyzed. Since our work utilizes FC-tweets, analyzing users' replies \cite{qazvinian2011rumor,friggeri2014rumor,vosoughi2018spread,jiang2018linguistic} are closely related to ours. However, the prior works had limited attention on analyzing how fact-checkers convey fact-checked content to original posters in public discourse. 

Additionally, researchers investigated topical interests and temporal behavior of fact-checkers \cite{vo2018rise}, relationship between fact-checkers and original posters \cite{hannak2014get}, how fake news disseminated when fact-checked evidences appeared \cite{friggeri2014rumor}, and whether users were aware of fact-checked information when it was available \cite{jiang2018linguistic}. Our work is different from these prior works since we focus on linguistic dimensions of FC-tweets, and propose and build a response generation framework to support fact-checkers.




\subsection{Applications of Text Generation}
Text generation has been used for language modeling \cite{mikolov2010recurrent}, question and answering \cite{hermann2015teaching}, machine translation \cite{sutskever2014sequence, bahdanau2014neural, luong2015effective}, dialogue generation \cite{serban2015building, shang2015neural,vinyals2015neural,serban2017hierarchical,wang2018chat}, and so on. Recently, it is employed to build chat bots for patients under depression\footnote{\url{https://read.bi/2QZ0ZPn}}, customer assistants in commercial sites, teen chat bots \cite{MicrosoftBot}, and supporting tools for teachers \cite{chen2018learningq}. Text generation has been also used to detect fake review \cite{yao2017automated}, clickbait headlines \cite{shu2018deep}, and fake news \cite{qian2018neural}. Our study is the first work that generates responses based on FC-tweets as a supporting tool for fact-checkers. Our work is closely related with dialog generation in which there are three main technical directions: (1) deterministic models \cite{serban2015building, shang2015neural,vinyals2015neural,wang2018chat}, (2) Variational Auto-Encoders (VAEs) \cite{serban2017hierarchical}, and (3) Generative Adversarial Networks (GANs) \cite{li2017adversarial}. Although recently VAEs and GANs showed promising results, deterministic models are still dominant in literature since they are easier to train than VAEs and GANs, and achieve competitive results \cite{le2018variational}. Thus, we propose a response generation framework based on Seq2Seq and attention mechanism \cite{luong2015effective}.

\section{Dataset}
\label{sec:data_collection}
In this section, we describe our data collection and preprocessing process. We utilized the dataset in \cite{vo2018rise} and the Hoaxy system \cite{shao2016hoaxy} to collect FC-tweets, which contained fact-checking URLs from two popular fact-checking sites: \textit{snopes.com}, and  \textit{politifact.com}. Totally, we collected 247,436 distinct fact-checking tweets posted between May 16, 2016 and May 26, 2018. 

Similar to \cite{vo2018rise, hannak2014get}, we removed non-English FC-tweets, and FC-tweets containing fact-checking URLs linked to non-article pages such as the main page and about page of a fact-checking site. Then, among the remaining fact-checking pages, if its corresponding original tweet was deleted or was not accessible via Twitter APIs because of suspension of an original poster, we further filtered out the fact-checking tweet. As a result, 190,158 FC-tweets and 164,477 distinct original tweets were remained.

To further ensure that each of the remaining FC-tweets reflected fact-checking intention and make a high quality dataset, we only kept a fact-checking tweet whose fact-checking article was rated as true or false. Our manual verification of 100 random samples confirmed that fact-checking tweets citing fact-checking articles with true or false label contained clearer fact-checking intention than fact-checking tweets with other labels such as half true or mixture. 
In other words, FC-tweets associated with mixed labels were discarded. After the pre-processing steps, our final dataset consisted of 73,203 FC-tweets and 64,110 original tweets posted by 41,732 distinct fact-checkers, and 44,411 distinct original posters, respectively. We use this dataset in the following sections.

\section{Linguistic Characteristics of Fact-checking Tweets}
\label{sec:data_analysis}
Since our goal is to automatically generate responses with fact-checking intention, it is necessary to analyze what kind of linguistic characteristics FC-tweets have, and verify whether FC-tweets in our dataset have the fact-checking intention.

To highlight linguistic characteristics of FC-tweets, we compare FC-tweets with \textit{Normal Replies}, which are direct responses to the same 64,110 original tweets without including fact-checking URLs, and \textit{Random Replies}, which do not share any common original tweets with FC-tweets. Initially, we collected 262,148 English \textit{Normal Replies} and 97M English \textit{Random Replies} posted in the same period of the FC-tweets. Then, we sampled 73,203 Normal Replies and 73,203 Random Replies from the initial collection to balance the data with our FC-tweets. All of the FC-tweets, Normal Replies and Random Replies were firstly preprocessed by replacing URLs with \textit{url} and mentions with \textit{@user}, and by removing special characters. They were tokenized by the NLTK tokenizer. Then, we answer the following research questions. Note that we sampled 73,203 Random Replies and 73,203 Normal Replies four times more, and our analysis was consistent with the following results.


\begin{table}[t]
	\caption{5 LDA topics and associated keywords of FC-tweets, Normal Replies and Random Replies.}
	\vspace{-5pt}
	\resizebox{1.0\linewidth}{!}
	{
	\begin{tabular}{l|lllll}
		\toprule[1pt] 
		Types                                                                      & Topic 1 & Topic 2 & Topic 3  & Topic 4   & Topic 5 \\ \hline
		\multirow{3}{*}{FC-tweets}                                                 & read    & fake    & facts    & false     & true    \\
		& stop    & news    & check    & snopes    & know    \\
		& try     & that's  & debunked & lie       & story   \\ \hline
		\multirow{3}{*}{\begin{tabular}[c]{@{}l@{}}Normal \\ Replies\end{tabular}} & one     & u       & fake     & president & like    \\
		& liar    & like    & news     & trump     & really  \\
		& know    & f***    & trump    & get       & stop    \\ \hline
		\multirow{3}{*}{\begin{tabular}[c]{@{}l@{}}Random\\ Replies\end{tabular}}  & love    & good    & like     & like      & thank   \\
		& u       & thanks  & one      & would     & oh      \\
		& know    & yes     & people   & right     & im      \\ \bottomrule[1pt] 
	\end{tabular}
	}
\label{tbl:lda_topics_dtweets}
\vspace{-10pt}
\end{table}

\smallskip
\noindent\textbf{Q1: What are underlying themes in FC-tweets?}

To answer this question, we applied the standard LDA algorithm to each of the three types of replies, so we built three independent LDA models. Table \ref{tbl:lda_topics_dtweets} shows 5 topics extracted from each of the three LDA topic models with associated keywords. Firstly, FC-tweets exhibit clear fact-checking intention with keywords such as debunked, snopes, read, stop, check, and lie. Secondly, keywords of Normal Replies show awareness of misinformation. However, fact-checking intention is not clear compared with FC-tweets. The keywords of Random Replies are commonly used in daily conversations. Based on the analysis, we conclude that the main themes of FC-tweets are about fact-checking information in the original tweets.

\begin{figure*}[t]
	\centering
	\subfigure[Pronouns]{
		\label{fig:pronouns_dtweets_random_third}
		\includegraphics[trim=0 0 50 50,clip,width=0.23\linewidth,height=1.4in]{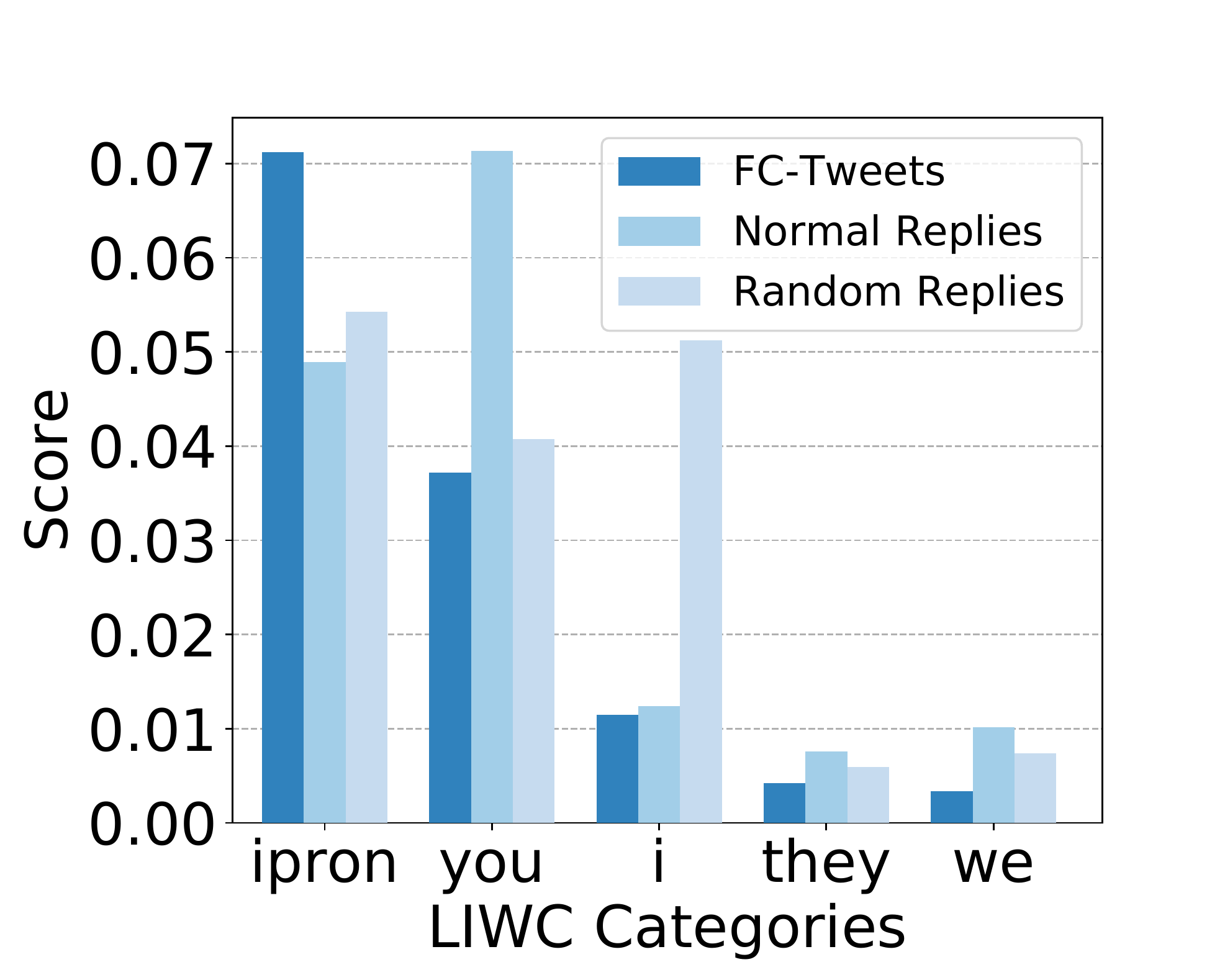}
	}
	\subfigure[Negation]{
		\label{fig:linguistics_dimensions_dtweets_random_third}
		\includegraphics[trim=5 5 40 40,clip,width=0.23\linewidth,height=1.4in]{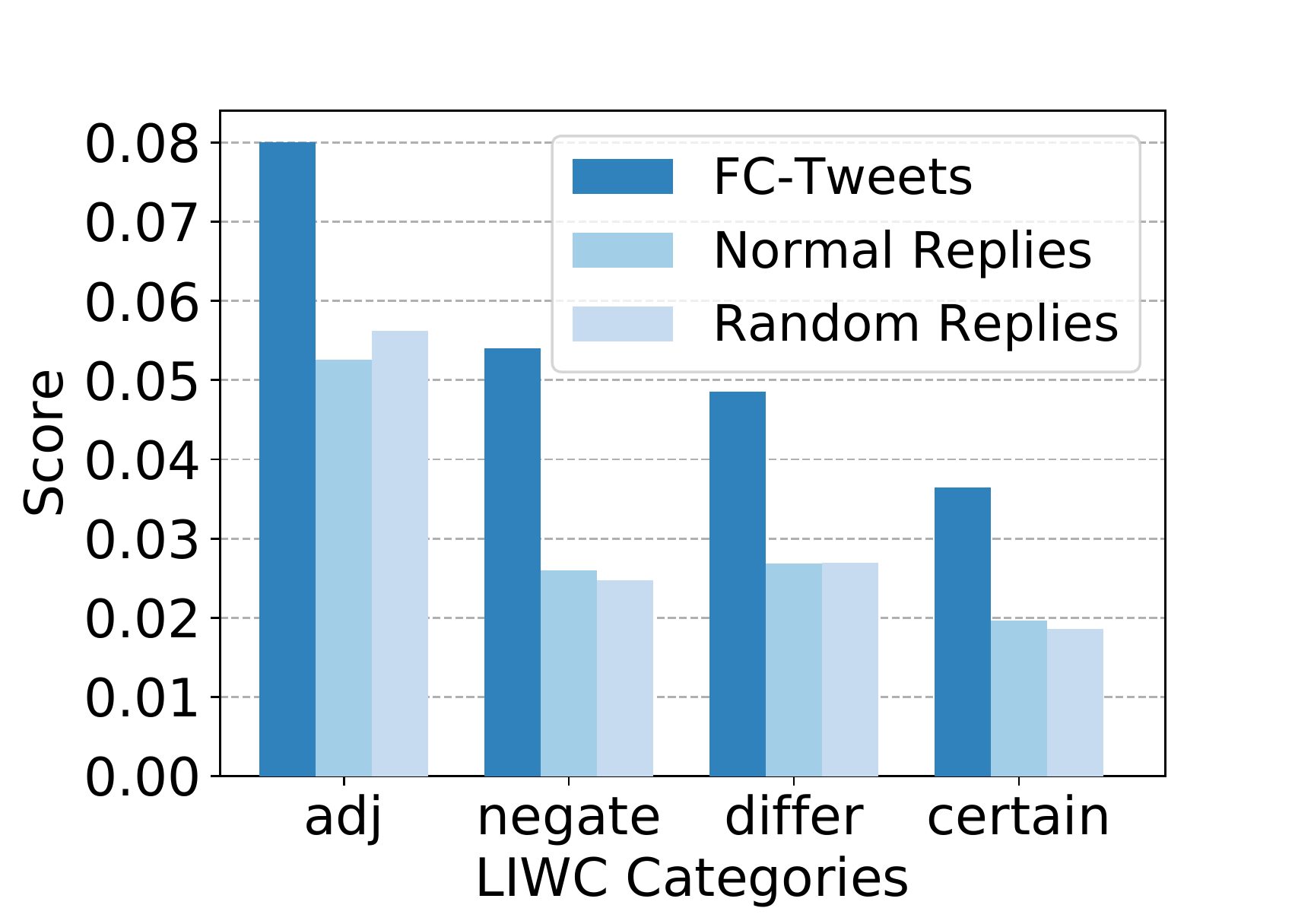}
	}
	\subfigure[Informality]{
		\label{fig:informal_languages_dtweets_random_third}
		\includegraphics[trim=5 5 50 40,clip,width=0.23\linewidth,height=1.4in]{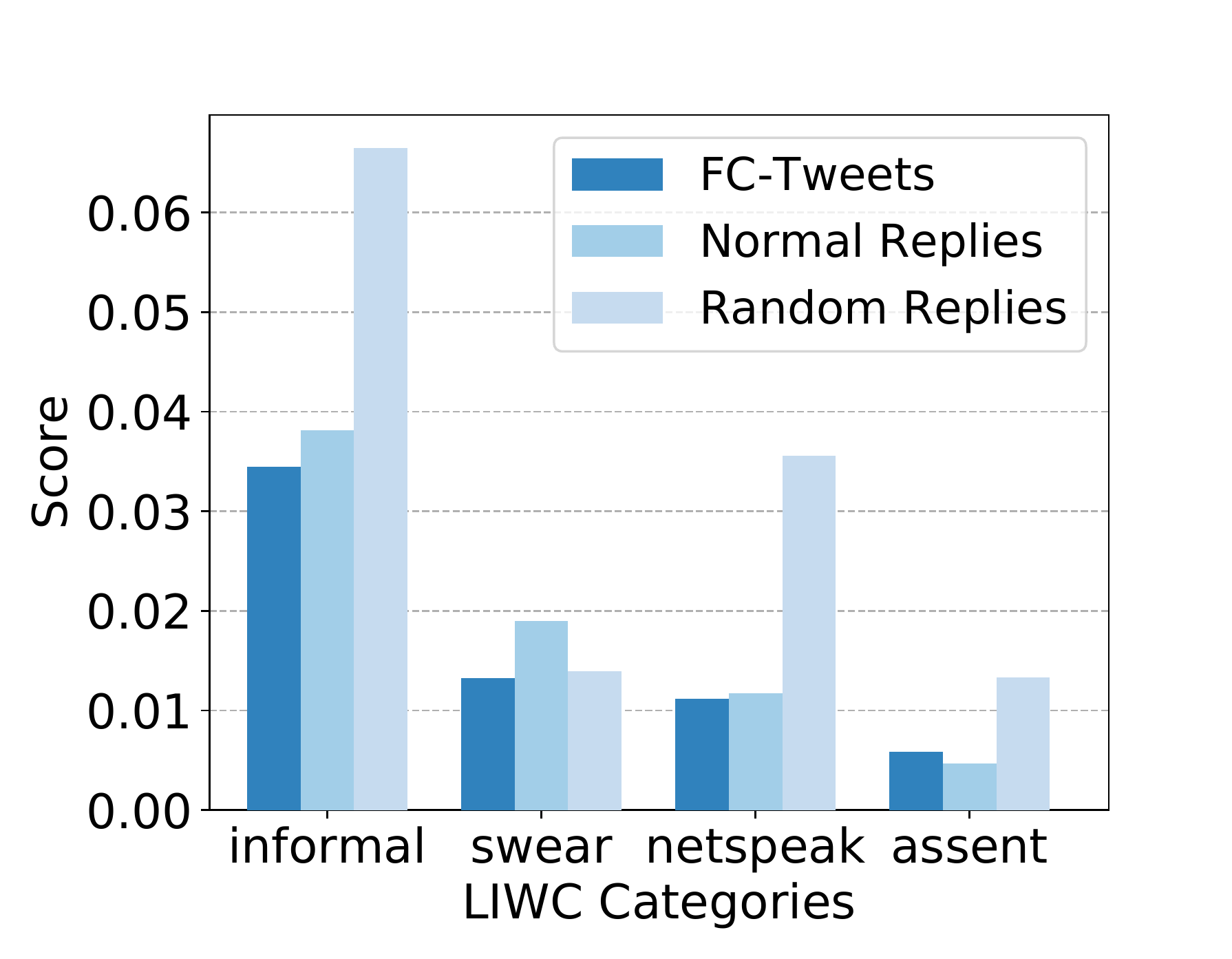}
	}
	\subfigure[Temporal Dimensions]{
		\label{fig:temporal_features_dtweets_random_third}
		\includegraphics[trim=5 5 50 40,clip,width=0.23\linewidth,height=1.4in]{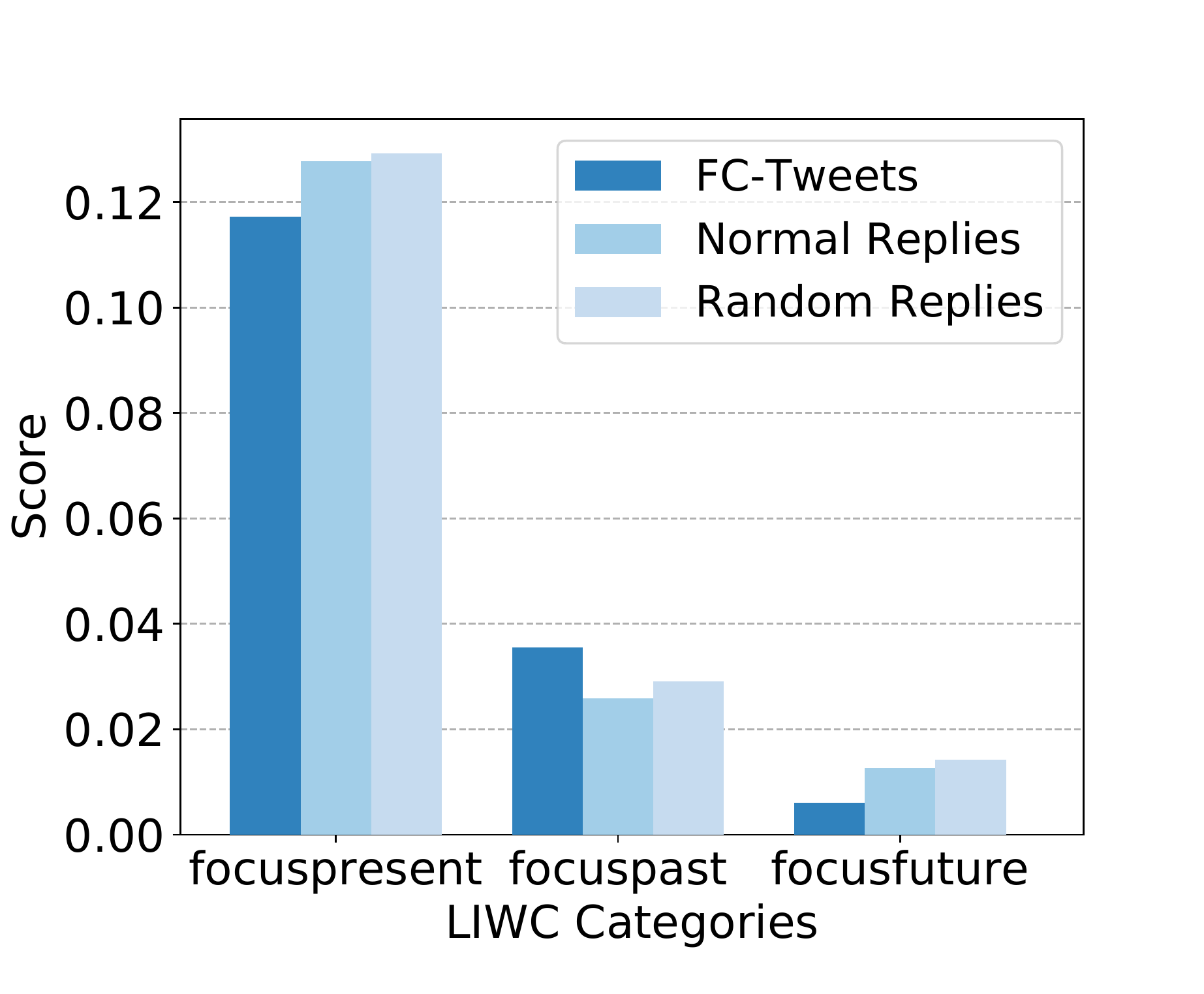}
	}
	\vspace{-5pt}
	\caption{LIWC category scores of FC-tweets, Normal Replies and Random Replies. In the figure, we follow LIWC category abbreviations as labels \cite{pennebaker2015development}. For example, \emph{ipron} and \textit{adj} mean impersonal pronouns and adjectives, respectively.}
	\vspace{-5pt}
\end{figure*}

\smallskip
\noindent\textbf{Q2: What are the psycholinguistic characteristics of FC-tweets?}

We employed LIWC 2015 \cite{pennebaker2015development}, a standard approach for mapping text to 73 psychologically-meaningful categories, for understanding psychological characteristics of FC-tweets. Given each of FC-tweets, we counted how many words of the FC-tweet belonged to each LIWC category. Then, we computed a normalized score for the category by diving the count by the number of words in the FC-tweet. Finally, we report the average scores $\mu$ and variances $\sigma^2$ for each LIWC category based on $|$FC-tweets$|$. The same process was applied to Normal Replies and Random Replies. We examined all LIWC categories and report only the most significant results. 
\noindent\textbf{(A) FC-tweets have the highest usage of impersonal pronouns and the least utilization of personal pronouns.} In Figure \ref{fig:pronouns_dtweets_random_third}, we can see that FC-tweets exhibit the highest usage of impersonal pronouns (e.g. it, this, that) ($\mu=0.071,\sigma^2=0.01$) in comparison with Normal Replies ($\mu=0.049,\sigma^2=0.009$) and Random Replies ($\mu=0.054,\sigma^2=0.005$). This observation is statistically significant in Mann Whitney one sided U-test ($p<0.001$). Examples of FC-tweets containing impersonal pronouns (called \emph{iprons} in LIWC) are (i) \textit{@user \textbf{This} has been debunked repeatedly - url please stop spreading the lie, thanks!}, and (ii) \textit{@user \textbf{it} is a wonderful quote but Lewis never said it : url and url}. 

Differently, Normal Replies show the highest mean score in 2nd person pronouns (named \textit{you} category) $(\mu=0.071,\sigma^2=0.007,p<0.001)$ in comparison to FC-tweets $(\mu=0.037)$ and Random Replies $(\mu=0.047)$. Note that \textit{you} in this context may refer to the original posters. In 1st person pronouns (named \textit{I} category), Random Replies have highest score because they contain daily personal conversations between online users. Finally, FC-tweets have the smallest usage of \textit{we} ($\mu=0.003,\sigma^2=0.000$) and \textit{they} ($\mu=0.004,\sigma^2=0.000$) among three groups of replies $(p<0.001)$. 


\noindent\textbf{(B) FC-tweets have a tendency to refute content of original tweets.}
Figure \ref{fig:linguistics_dimensions_dtweets_random_third} shows the mean scores of \textit{adj}, \textit{negate}, \textit{differ} and \textit{certain} categories. Specifically, FC-tweets exhibit the highest mean score in \textit{adjectives} category ($\mu=0.080,\sigma^2=0.024,p<0.001$) in comparison to Normal Replies ($\mu=0.052,\sigma^2=0.008$) and Random Replies $(\mu=0.056,\sigma^2=0.015)$. Prevalent adjectives in FC-tweets are fake, wrong, dump, false, and untrue. FC-tweets also tend to refute information of original tweets. Their mean score in \textit{negate} category is $0.054$, which is about two times higher than the mean score of Normal Replies $(p<0.001)$. 
FC-tweets have also the highest usage of words in \textit{differ} category (e.g. actually, but, except) among the three groups $(p<0.001)$. In \textit{certain} category (e.g. never, ever, nothing, always), FC-tweets' mean score $(\mu=0.036)$ also doubles the average score of Normal Replies significantly $(p<0.001)$. Examples of FC-tweets are: (i) \textit{@user \textbf{wrong}. \textbf{never} happened. url}, (ii) \textit{@user except he \textbf{didn't}. that tweet has been proven \textbf{fake}: url}, and (iii) \textit{@user I \textbf{sure} hope you're joking. url}.

\noindent\textbf{(C) FC-tweets are usually more formal and have low usage of swear words.} In Figure \ref{fig:informal_languages_dtweets_random_third}, FC-tweets have lower mean score in  \textit{informal} category $(\mu=0.034,\sigma^2=0.010)$ than Normal Replies $(\mu=0.038, p<0.001)$ and Random Replies $(\mu=0.066)$. FC-tweets also use the least swear words $(\mu=0.013,\sigma^2=0.005, p<0.005)$ among the three groups. In terms of \textit{netspeak} category (i.e. Internet slangs), FC-tweets ($\mu\approx0.011$) generally have smaller average score than Random Replies $(\mu=0.035)$. Furthermore, FC-tweets do not contain much words in \textit{assent} category (e.g. OK, yup, okey) ($\mu=0.006,\sigma^2=0.002$) compared with Random Replies $(\mu=0.013,\sigma^2=0.005,p<0.001)$. Regarding formality of FC-tweets, we conjecture that fact-checkers try to persuade original posters to stop spreading fake news, leading to more formal language, less usage of swear words. An example of FC-tweets is \textit{@user url I'm sure you'll still say it's true- but it simply isn't. Google for facts and debunks please}. 

\noindent\textbf{(D) FC-tweets emphasize on what happened in the past whereas Normal Replies and Random Replies focus on present and future.} In Figure \ref{fig:temporal_features_dtweets_random_third},
FC-tweets usually employ verbs in past tense to mention past stories to support their factual corrections. Thus, the average score of \textit{focuspast} category of FC-tweets is the highest $(\mu=0.036,\sigma^2=0.005,p<0.001)$ among the three groups of replies, whereas Normal Replies and Random Replies emphasize on present and future. Particularly, FC-tweets have the least score in \textit{focuspresent} $(\mu=0.117,\sigma^2=0.015,p<0.001)$ and \textit{focusfuture} $(\mu=0.006,\sigma^2=0.000,p<0.001)$ categories.
An example of FC-tweets is \textit{@user yeah, she merely \textbf{said} something that \textbf{was} only slightly less absurd. url}. 
\begin{figure}[t]
	\centering
	\includegraphics[width=0.9\linewidth,height=1.8in,trim=40 5 60 40,clip]{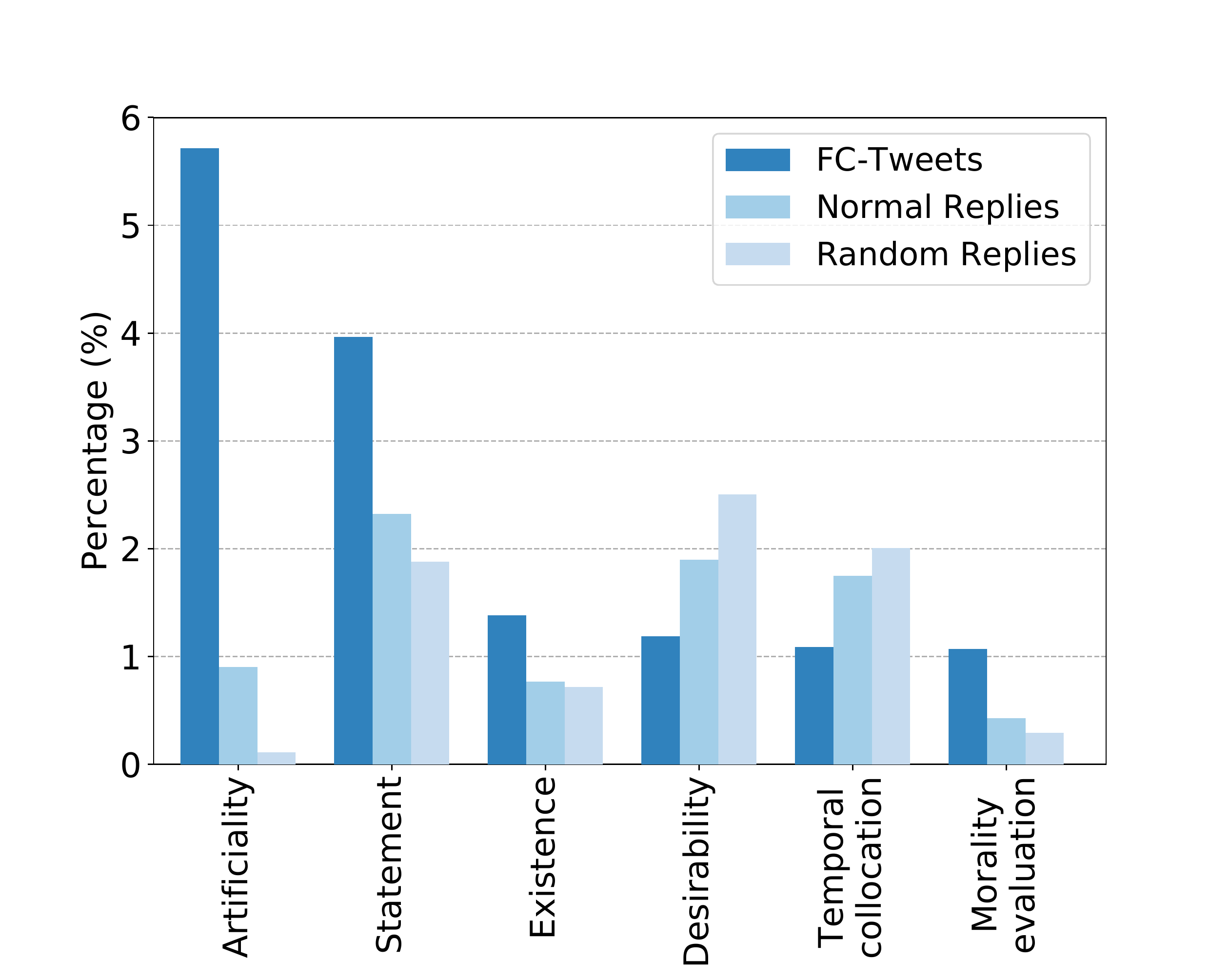}
	\caption{Distribution of semantic frame types in FC-tweets.}
	\label{fig:frame_net_distribution_dtweet_vs_other_replies_same_target_vs_random}
	\vspace{-10pt}
\end{figure}

\smallskip
\noindent\textbf{Q3: How are semantic frames represented in FC-tweets?}

So far, we examined every word independently without considering its dependencies with other words (e.g. surrounding words), which is helpful in understanding its meaning. Thus, we now employ SEMAFOR model \cite{das2010probabilistic}, trained on FrameNet data\footnote{\url{https://framenet.icsi.berkeley.edu/fndrupal/frameIndex}}, to extract rich structures called \textit{semantic frames} based on syntactic trees of sentences. For example, a frame \textit{Statement} consists of \textit{a noun} and \textit{a verb} where \textit{the noun} indicates a speaker and \textit{the verb} implies the act of conveying a message. We measured the distribution of semantic frames of FC-tweets by firstly counting the number of occurrences of every frame type across all FC-tweets, and normalized it by the total number of detected frames in all FC-tweets. The same process was applied to Normal Replies and Random Replies. Figure \ref{fig:frame_net_distribution_dtweet_vs_other_replies_same_target_vs_random} shows the percentage of different types of frames detected by SEMAFOR. We have the following observations:

\noindent\textbf{(A) FC-tweets display high usage of \textit{Artificiality}, \textit{Statement} and \textit{Existence}.} In Figure \ref{fig:frame_net_distribution_dtweet_vs_other_replies_same_target_vs_random}, FC-tweets have the highest utilization of \textit{Artificiality} (e.g. wrong, lie, fake, false, genuine, phoney) among three groups of replies ($p<0.001$ according to one-sided z-test). This frame accounts for 5.71\% detected frames in FC-tweets compared with Normal Replies (0.90\%) and Random Replies (0.11\%). FC-tweets also have the highest proportion of the frame \textit{Statement} $(3.96\%, p<0.001)$ among three groups of replies. Words that evoke frame \textit{Statement} in FC-tweets are said, says, claims, report, told, talk, and mention. Examples of FC-tweets are (i) \textit{@user You're the one who has no clue. She never \textbf{said} this: url}, and (ii) \textit{@user Snopes \textbf{reports} this rumor as \textbf{false}. url}. 
To refer to verified information, FC-tweets employed frame \textit{Existence} (1.38\%, $p<0.001$) compared with Normal Replies (0.71\%) and Random Replies (0.70\%). The most popular phrases evoking frame \textit{Existence} were real, there is, there are, exist, there were, and there have been. Examples of FC-tweets are: (i) \textit{@user \textbf{There is} no trucker strike in Puerto Rico url}, (ii) \textit{@user That town doesnt \textbf{exist} url}. 

\noindent\textbf{(B) FC-tweets exhibit the highest \textit{Morality\_evaluation} and have less usage of \textit{Desirability}.} As shown in Figure \ref{fig:frame_net_distribution_dtweet_vs_other_replies_same_target_vs_random}, FC-tweets contain the highest proportion of frame \textit{Morality\_evaluation} (1.06\%, $p<0.001$) among three groups. The most popular words in frame \textit{Morality\_evaluation} are wrong, evil, dishonest, despicable, unethical, and immoral. Another supporting evidence of this observation is the lower usage of frame \textit{Desirability} (e.g. good, better, bad, great, okay, cool) in FC-tweets (1.19\%, $p<0.001$) than Normal Replies (1.89\%) and Random Replies (2.5\%). An example of FC-tweets is: \textit{@user you're such an \textbf{evil}, \textbf{despicable} creature. url}

\noindent\textbf{(C) FC-tweets do not use much \textit{Temporal\_collocation}.}
FC-tweets show lower usage of \textit{Temporal\_collocation} (1.08\%, $p<0.001$) than Normal Replies $(1.74\%)$ and Random Replies $(2.00\%)$. The most common words that evoke this frame in FC-tweets are when, now, then, today, current, recently, future. It seems these words are mainly about the present and the future. This result supports the same observation that FC-tweets tend to focus on the past. 

\begin{figure}[t]
	\centering
	\subfigure[FC-tweets vs. articles]{
		\label{fig:cosine_sim_boxplot}
		\includegraphics[trim=5 0 40 40,clip,width=0.3\linewidth,height=1in]{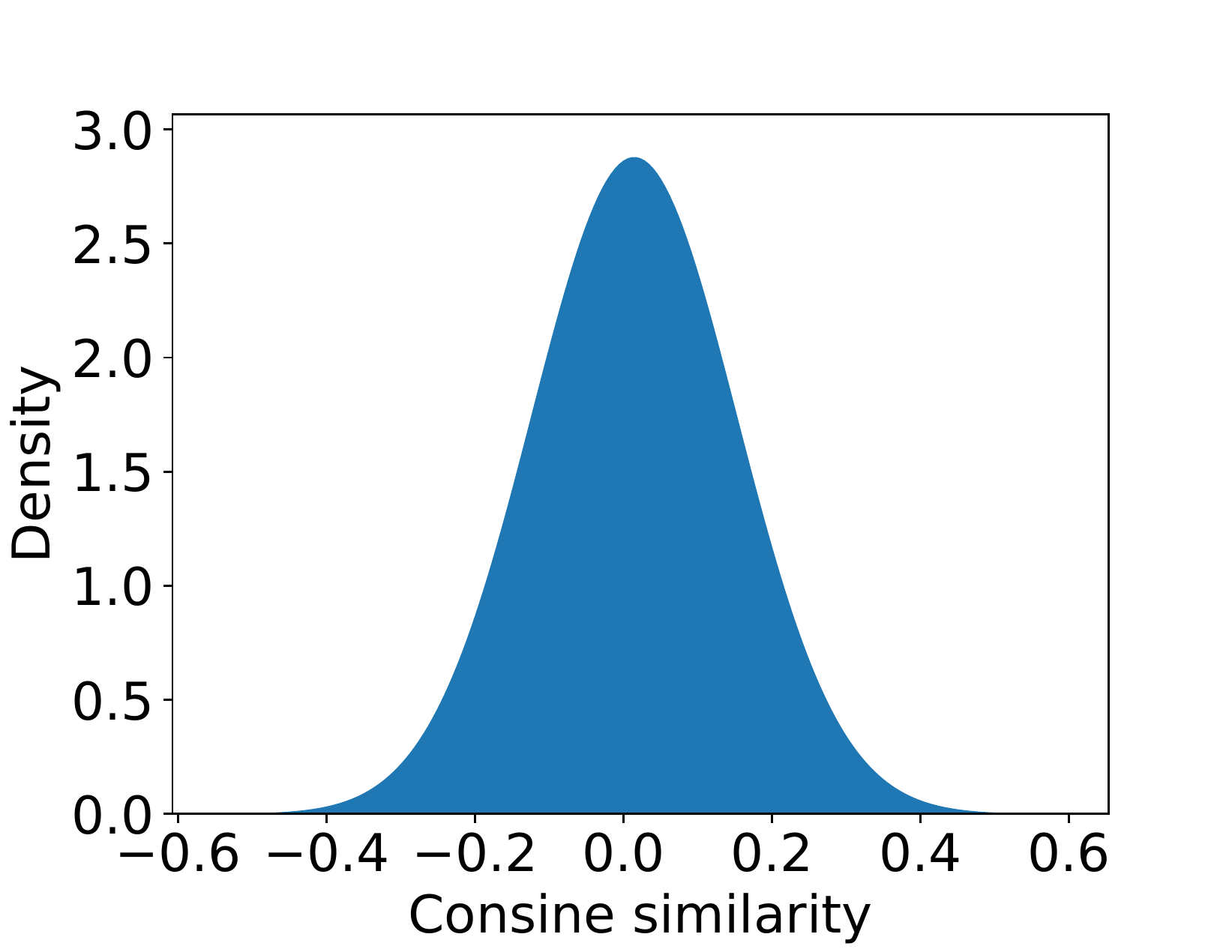}
	}
	\subfigure[$|$Shares$|$ vs. $|$tokens$|$]{
		\label{fig:length_vs_shares}
		\includegraphics[trim=0 5 40 35,clip,width=0.3\linewidth,height=1in]{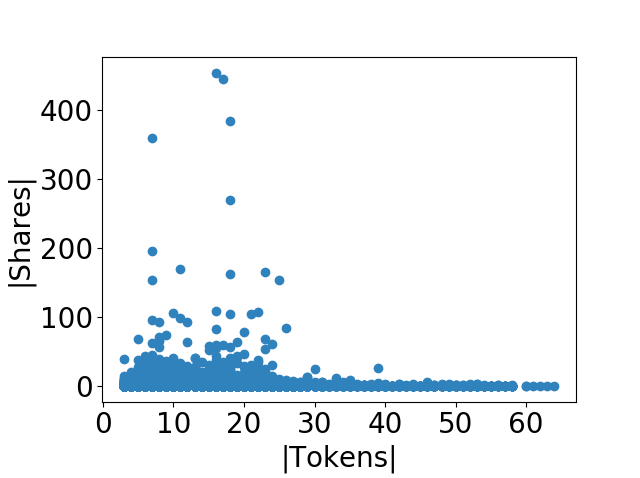}
	}
	\subfigure[Irretrievable tweets]{
		\label{fig:pie_chart_deleted_status}
		\includegraphics[trim=0 0 35 35,clip,width=0.3\linewidth,height=1in]{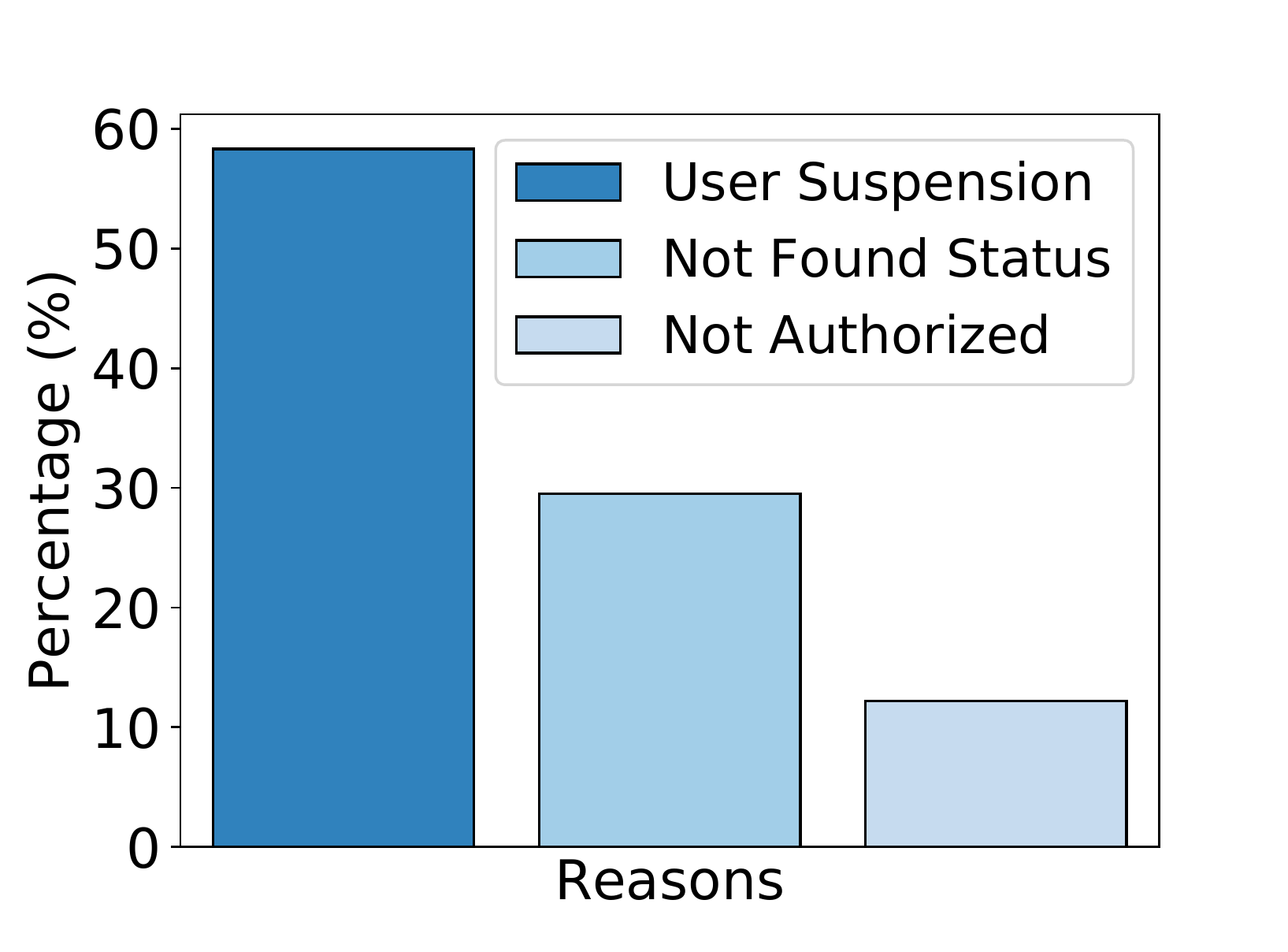}
	}
	\vspace{-5pt}
	\caption{(a) Similarity between FC-tweets and fact-checking articles, (b) $|$Shares$|$ vs. $|$Tokens$|$ of FC-tweets and (c) Distribution of irretrievable original tweets. 
	}
	\vspace{-10pt}
\end{figure}

\smallskip
\noindent\textbf{Q4: Do fact-checkers include details of fact-checking articles?}
We firstly derived latent representations of FC-tweets and articles by training two Doc2Vec models \cite{le2014distributed} - one for FC-tweets and the other one for fact-checking articles. The embedding size is 50. Then, we measured cosine similarity between a FC-tweet and the fact-checking article embedded in the FC-tweet as shown in Figure \ref{fig:cosine_sim_boxplot}. 
Interestingly, most FC-tweets do not have high similarity with FC-articles, suggesting that fact-checkers rarely include details from fact-checking articles in FC-tweets. However, there were several enthusiastic fact-checkers who extracted information from fact-checking articles to make FC-tweets more persuasive, as shown in two tails of the curve in Figure \ref{fig:cosine_sim_boxplot}.

\smallskip
\noindent\textbf{Q5: Is there any connection between $|$tokens$|$ of FC-tweets and $|$shares$|$?}
Since sharing FC-tweets by retweets and quotes is important for increasing the visibility of credible information on online social networks, we examined correlation between $|$tokens$|$ of FC-tweets and their $|$shares$|$. We only focus on $|$tokens$|$ because it could help us to decide length of a generated response.
Figure \ref{fig:length_vs_shares} shows a scatter plot of FC-tweets' $|$tokens$|$ and $|$shares$|$ (i.e., quotes and retweets). Generally, most FC-tweets had $|$shares$|$=0. However, FC-tweets with $|$tokens$|$ $\in[10;20]$ usually received more attention. To verify this, we created two lists -- one containing $|$shares$|$ of FC-tweets with $|$tokens$|$ $\in[0;9]$ and another one for $|$shares$|$ of FC-tweets with $|$tokens$|$ $\in[10;20]$ --, and then conducted Mann Whitney one-sided U-test. We found that the latter one had significantly larger numbers than the former one $(p=2.91\times 10^{-11})$. We conclude that very short FC-tweets may be not informative enough to draw readers' attention, and lengthy FC-tweets may be too time-consuming to read, leading to small number of shares. Therefore, a reasonable length of FC-tweets is more preferable when we generate a response.




\smallskip
\noindent\textbf{Q6: Is there any signal suggesting positive effect of FC-tweets?}
We examined what happened to original tweets after receiving fact-checking tweets. In Oct 2018 (i.e., five months after collecting our dataset), we re-collected original tweets via Twitter APIs to see if all of the original tweets were retrievable. Interestingly, 4,516 (7\%) original tweets were not retrievable. There are three reasons: (i) \textit{User Suspension}, (ii) \textit{Not Found Status} (i.e. deleted status), and (iii) \textit{Not Authorized} (i.e. original tweets are in the private mode).

In Figure \ref{fig:pie_chart_deleted_status}, \textit{User Suspension} accounted for 58.30\% of the irretrievable original tweets. Although there may be many factors that potentially explain suspension (e.g. original posters may have other abusing behaviors that triggered Twitter security system), one obvious observation is that fact-checkers tended to target bad users (e.g. content polluters \cite{lee2010uncovering}), who usually have abusing behavior on social platform. It means that fact-checkers are enthusiastic about checking credibility of information on social networks. Regarding two reasons \textit{Not Authorized} and \textit{Not Found Status}, perhaps original posters were either aware of the wrong information they posted or were under pressure due to criticisms they received from other users, leading to deletion or hiding their original tweets.

In summary, our analysis reveals fact-checkers refuted content of original tweets, and their FC-tweets were more formal than Normal Replies and Random Replies. To provide supporting evidences, FC-tweets utilized semantic frames \emph{Existence} and \emph{Statement}. These results confirm that FC-tweets exhibit clear fact-checking intention.


\section{Response Generation Framework}
In the previous section, we analyzed common topics, lexical usages, and distinguishing linguistic dimensions of FC-tweets compared with Normal Replies and Random Replies. Our analysis revealed that FC-tweets indeed exhibited clear fact-checking intention, which is the property that we desired. Now, we turn our attention to proposing and building our framework, named \emph{Fact-checking Response Generator} (\textbf{FCRG}), in order to generate responses with fact-checking intention. The generated responses are used to support fact-checkers and increase their engagement. 

Formally, given a pair of an original tweet and a FC-tweet, the original tweet $x$ is a sequence of words $x = \{x_i|i\in [1;N]\}$ and the FC-tweet is another sequence of words $y = \{y_j|j\in [1;M]\}$, where $N$ and $M$ are the length of the original tweet and the length of FC-tweet, respectively. We inserted a special token $\texttt{<s>}$ as a starting token into every FC-tweet. Drawing inspiration from \cite{luong2015effective}, we propose and build a framework as shown in Figure \ref{fig:framework_big_picture} that consists of three main components: (i) the shared word embedding layer, (ii) the encoder to capture representation of the original tweet and (iii) the decoder to generate a FC-tweet. Their details are as follows:

\subsection{Shared Word Embedding Layer}
For every word $x_i$ in the original tweet $x$, we represent it as a one-hot encoding vector $x_i \in \mathbb{R}^V$ and embed it into a \textit{D}-dimensional vector $\textbf{x}_i \in \mathbb{R}^{D}$ as follows: $\textbf{x}_i = \textbf{W}_ex_i$, where $\textbf{W}_e \in \mathbb{R}^{D\times V}$ is an embedding matrix and $V$ is the vocabulary size. We use the same word embedding matrix $\textbf{W}_e$ for the FC-tweet. In particular, for every word $y_i$ (represented as one-hot vector $y_i \in \mathbb{R}^V$) in the FC-tweet $y$, we embed it into a vector $\textbf{y}_i = \textbf{W}_ey_i$. The embedding matrix $\textbf{W}_e$ is a learned parameter and could be initialized by either pre-trained word vectors (e.g. Glove vectors) or random initialization. Since our model is designed specifically for fact-checking domain, we initialized $\textbf{W}_e$ with Normal Distribution $\mathcal{N}(0, 1)$ and trained it from scratch. By using a shared $\textbf{W}_e$, we could reduce the number of learned parameters significantly compared with \cite{luong2015effective}. This is helpful in reducing overfitting.

\subsection{Encoder}
The encoder is used to learn latent representation of the original tweet $x$. We adopt a Recurrent Neural Network (RNN) to represent the encoder due to its large capacity to condition each word $x_i$ on all previous words $x_{<i}$ in the original tweet $x$. To overcome the vanishing or exploding gradient problem of RNN, we choose Gated Recurrent Unit (GRU) \cite{cho2014learning}. Formally, we compute hidden state $\textbf{h}_i \in \mathbb{R}^{H}$ at time-step $i^{th}$ in the encoder as follows:
\begin{equation}
	\textbf{h}_i = GRU(\textbf{x}_i, \textbf{h}_{i-1})
	\label{eq:GRU}
\end{equation}
where the GRU is defined by the following equations:
\begin{equation}
	\begin{split}
	\textbf{z}_i &= \sigma(\textbf{x}_i\textbf{W}_z   + \textbf{h}_{i-1} \textbf{U}_z ) \\
	\textbf{r}_i &= \sigma(\textbf{x}_i\textbf{W}_r  +  \textbf{h}_{i-1}\textbf{U}_r \big) \\
	\tilde{\textbf{h}}_i &= tanh\big(\textbf{x}_i\textbf{W}_o  + (\textbf{r}_i \odot \textbf{h}_{i-1} )\textbf{U}_o) \\
	\textbf{h}_i &= (1 - \textbf{z}_i) \odot \tilde{\textbf{h}}_i + \textbf{z}_i \odot \textbf{h}_{i-1}
	\end{split}
\end{equation}
where $\textbf{W}_{[z,r,o]}, \textbf{U}_{[z,r,o]}$ are learned parameters. $\tilde{\textbf{h}}_i$ is the new updated hidden state, $\textbf{z}_i$ is the update gate, $\textbf{r}_i$ is the reset gate, $\sigma(.)$ is the sigmoid function, $\odot$ is element wise product, and $\textbf{h}_0 = \textbf{0}$. After going through every word of the original tweet $x$, we have hidden states for every time-step $\textbf{X} = [\textbf{h}_1 \oplus \textbf{h}_2\oplus...\oplus\textbf{h}_N] \in \mathbb{R}^{H\times N}$, where $\oplus$ denotes concatenation of hidden states. We use the last hidden state $\textbf{h}_N$ as features of the original tweet $\textbf{x} = \textbf{h}_N$.

\begin{figure}[t]
	\centering
	\includegraphics[trim=80 110 310 60,clip,width=\linewidth,height=2.3in]{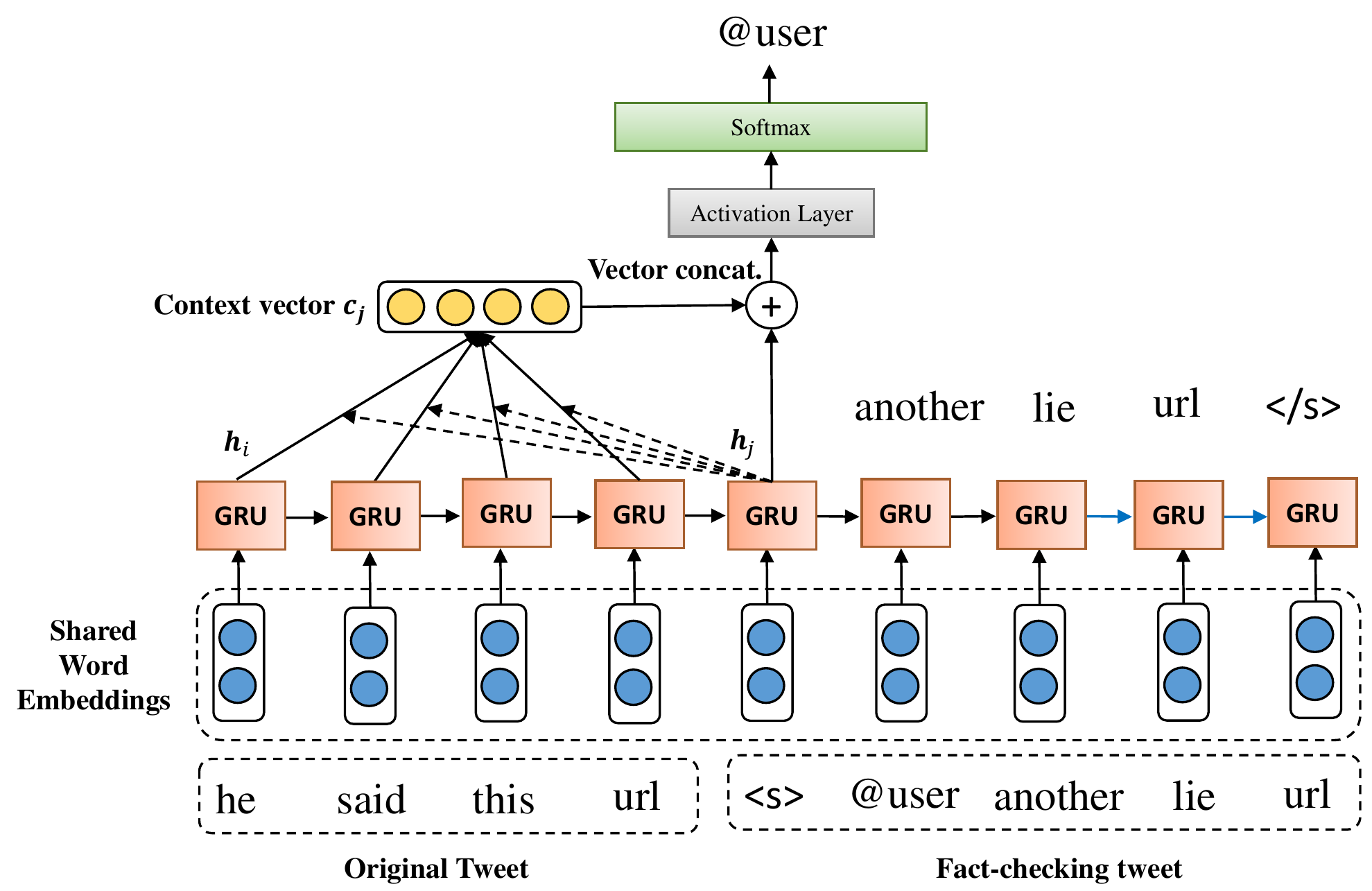}
	\caption{Our proposed framework to generate responses with fact-checking intention.}
	\label{fig:framework_big_picture}
\end{figure}

\subsection{Decoder}
The decoder takes $\textbf{x}$ as the input to start the generation of a FC-tweet. We use another GRU to represent the decoder to generate a sequence of tokens $y = \{y_1, y_2,..,y_M\}$. At each time-step $j^{th}$, the hidden state $h_j$ is computed by another GRU: $\textbf{h}_j = GRU(\textbf{y}_j, \textbf{h}_{j-1})$ where initial hidden states are $\textbf{h}_{0} = \textbf{x}$. To provide additional context information when generating word $y_j$, we apply an attention mechanism to learn a weighted interpolation context vector $\textbf{c}_j$ dependent on all of the hidden states output from all time-steps of the encoder. We compute $\textbf{c}_j = \textbf{X} \textbf{a}_j$  
where each component $\textbf{a}_{ji}$ of $\textbf{a}_j \in \mathbb{R}^N$ is the alignment score between the $j^{th}$ word in the FC-tweet and the $i^{th}$ output from the encoder. In this study, $\textbf{a}_{j}$ is computed by one of the following ways:

\begin{equation}
	\textbf{a}_{j} = \begin{cases}
	softmax(\textbf{X}^T\textbf{h}_{j}) & \text{Dot Attention}\\
	softmax(\textbf{X}^T\textbf{W}_a\textbf{h}_{j}) & \text{Bilinear Attention}\\
	\end{cases}
	\label{eq:attention_mechanism}
\end{equation}
where softmax(.) is a softmax activation function and $\textbf{W}_a \in \mathbb{R}^{H\times H}$ is a learned weight matrix. Note that we tried to employ other attention mechanisms including additive attention \cite{bahdanau2014neural} and concat attention \cite{luong2015effective} but the above attention mechanisms in Eq. \ref{eq:attention_mechanism} produced better results. After computing the context vector $\textbf{c}_j$, we concatenate $\textbf{h}^T_j$ with $\textbf{c}^T_j$ to obtain a richer representation. The word at $j^{th}$ time-step is predicted by a softmax classifier:
\begin{equation}
\begin{split}
\hat{\textbf{y}}_j &= softmax\big(\textbf{W}_s\tanh\big(\textbf{W}_c[\textbf{c}^T_j\oplus\textbf{h}^T_j]^T\big)\big)
\end{split}
\label{eq:softmax_layer}
\end{equation}
where $\textbf{W}_c \in \mathbb{R}^{O\times 2H}$, and $\textbf{W}_s \in \mathbb{R}^{V\times O}$ are weight matrices of a two-layer feedforward neural network and $O$ is the output size. $\hat{\textbf{y}}_j \in \mathbb{R}^V$ is a probability distribution over the vocabulary. The probability of choosing word $v_k$ in the vocabulary as output is:
\begin{equation}
	p(y_j=v_k|y_{j-1},y_{j-2},...,y_1,\textbf{x}) = \hat{\textbf{y}}_{jk}
\end{equation}
Therefore, the overall probability of generating the FC-tweet $y$ given the original tweet $x$ is computed as follows:
\begin{equation}
	p(y|x) = \prod_{j=1}^{M} p(y_j|y_{j-1},y_{j-2},...,y_1,\textbf{x})
\end{equation}
Since the entire architecture is differentiable, we jointly train the whole network with Teacher Forcing via Adam optimizer \cite{kingma2014adam} by minimizing the negative conditional log-likelihood for $m$ pairs of the original tweet $x^{(i)}$ and the FC-tweet $y^{(i)}$ as follows:
\begin{equation}
	\min_{\theta_e,\theta_d}\mathcal{L} = -\sum_{i=1}^{m} \log{p(y^{(i)}|x^{(i)};\theta_e,\theta_d)}
\end{equation}
where $\theta_e$ and $\theta_d$ are the parameters of the encoder and the decoder, respectively. At test time, we used beam search to select top K generated responses. The generation process of a FC-tweet is ended when an end-of-sentence token (e.g. $\texttt{</s>}$) is emitted. 


\begin{table*}[]
	\caption{Performance of our models and baselines.}
	\vspace{-5pt}
	\begin{tabular}{l|c|lllll|cc|c}
		\hline
		Constraints                                                                     & Model    & BLEU-2    & BLEU-3    & BLEU-4    & ROUGE-L    & METEOR     & Greedy Mat.    & Vector Ext.   & Avg. Rank \\ \hline
		\multirow{4}{*}{\begin{tabular}[c]{@{}l@{}}Without \\ Constraints\end{tabular}} & SeqAttB  & 7.148 \textcolor{blue}{(4)} & 4.050 \textcolor{blue}{(4)} & 3.261 \textcolor{blue}{(3)} & 26.474 \textcolor{blue}{(3)} & 17.659 \textcolor{blue}{(3)} & 43.566 \textcolor{blue}{(3)} & 15.837 \textcolor{blue}{(4)} & 3.43      \\
		& HRED     & 7.301 \textcolor{blue}{(3)} & 4.073 \textcolor{blue}{(3)} & 3.248 \textcolor{blue}{(4)} & 26.222 \textcolor{blue}{(4)} & 17.545 \textcolor{blue}{(4)} & 42.734 \textcolor{blue}{(4)} & 18.929 \textcolor{blue}{(3)} & 3.57      \\
		& our FCRG-BL & \textbf{7.678} \textcolor{blue}{(1)} & 4.270 \textcolor{blue}{(2)} & 3.406 \textcolor{blue}{(2)} & 27.142 \textcolor{blue}{(2)} & \textbf{17.871} \textcolor{blue}{(1)} & 43.714 \textcolor{blue}{(2)} & \textbf{20.244} \textcolor{blue}{(1)} & 1.57      \\
		& our FCRG-DT     & 7.641 \textcolor{blue}{(2)} & \textbf{4.303} \textcolor{blue}{(1)} & \textbf{3.500} \textcolor{blue}{(1)} & \textbf{27.352} \textcolor{blue}{(1)} & 17.750 \textcolor{blue}{(2)} & \textbf{45.302} \textcolor{blue}{(1)} & 19.993 \textcolor{blue}{(2)} & 1.43      \\ \hline
		\multirow{4}{*}{\begin{tabular}[c]{@{}l@{}}At least \\ 5 tokens\end{tabular}}   & SeqAttB  & 7.470 \textcolor{blue}{(4)} & 4.085 \textcolor{blue}{(4)} & 3.175 \textcolor{blue}{(4)} & 26.169 \textcolor{blue}{(4)} & 17.719 \textcolor{blue}{(3)} & 41.038 \textcolor{blue}{(4)} & 14.686 \textcolor{blue}{(4)} & 3.86      \\
		& HRED     & 7.631 \textcolor{blue}{(3)} & 4.155 \textcolor{blue}{(3)} & 3.227 \textcolor{blue}{(3)} & 26.241 \textcolor{blue}{(3)} & 17.617 \textcolor{blue}{(4)} & 41.930 \textcolor{blue}{(3)} & 18.850 \textcolor{blue}{(3)} & 3.14      \\
		& our FCRG-BL & 7.925 \textcolor{blue}{(2)} & 4.285 \textcolor{blue}{(2)} & 3.295 \textcolor{blue}{(2)} & 26.953 \textcolor{blue}{(2)} & \textbf{17.885} \textcolor{blue}{(1)} & 42.899 \textcolor{blue}{(2)} & \textbf{20.052} \textcolor{blue}{(1)} & 1.71      \\
		& our FCRG-DT     & \textbf{8.043} \textcolor{blue}{(1)} & \textbf{4.373} \textcolor{blue}{(1)} & \textbf{3.409} \textcolor{blue}{(1)} & \textbf{27.020} \textcolor{blue}{(1)} & 17.770 \textcolor{blue}{(2)} & \textbf{44.379} \textcolor{blue}{(1)} & 19.441 \textcolor{blue}{(2)} & 1.29      \\ \hline
		\multirow{4}{*}{\begin{tabular}[c]{@{}l@{}}At least\\ 10 tokens\end{tabular}}   & SeqAttB  & 6.398 \textcolor{blue}{(4)} & 3.319 \textcolor{blue}{(4)} & 2.434 \textcolor{blue}{(4)} & 22.250 \textcolor{blue}{(4)} & 16.568 \textcolor{blue}{(4)} & 36.298 \textcolor{blue}{(4)} & 10.198 \textcolor{blue}{(4)} & 4.00      \\
		& HRED     & 6.540 \textcolor{blue}{(3)} & 3.373 \textcolor{blue}{(3)} & 2.462 \textcolor{blue}{(3)} & 22.980 \textcolor{blue}{(3)} & 17.106 \textcolor{blue}{(3)} & 37.513 \textcolor{blue}{(3)} & 15.537 \textcolor{blue}{(3)} & 3.00      \\
		& our FCRG-BL & 7.576 \textcolor{blue}{(2)} & 3.780 \textcolor{blue}{(2)} & 2.660 \textcolor{blue}{(2)} & \textbf{25.086} \textcolor{blue}{(1)} & \textbf{17.832} \textcolor{blue}{(1)} & \textbf{39.809} \textcolor{blue}{(1)} & \textbf{17.605} \textcolor{blue}{(1)} & 1.43      \\
		& our FCRG-DT     & \textbf{7.955} \textcolor{blue}{(1)} & \textbf{3.914} \textcolor{blue}{(1)} & \textbf{2.751} \textcolor{blue}{(1)} & 24.635 \textcolor{blue}{(2)} & 17.662 \textcolor{blue}{(2)} & 39.374 \textcolor{blue}{(2)} & 16.081 \textcolor{blue}{(2)} & 1.57      \\ \hline
	\end{tabular}
\label{tbl:performance_of_models_with_word_overlap_metrics}
\end{table*}

\section{Evaluation}
In this section, we thoroughly evaluate our models namely \textbf{FCRG-DT} (based on dot attention in Eq. \ref{eq:attention_mechanism}) and \textbf{FCRG-BL} (based on bilinear attention in Eq. \ref{eq:attention_mechanism}) quantitatively and qualitatively. 
We seek to answer the following questions:
\squishlist
\item \textbf{RQ1:} What are the performance of our models and baselines in word overlap-based metrics (i.e., measuring syntactic similarity between a ground-truth FC-tweet and a generated one)?
\item \textbf{RQ2: } How do our models perform compared with baselines in embedding metrics (i.e., measuring semantic similarity between a ground-truth FC-tweet and a generated one)?
\item \textbf{RQ3:} How does the number of generated tokens of responses affect performance of our models?
\item \textbf{RQ4:} Is our generated responses better than ones generated from baselines in a qualitative evaluation?
\item \textbf{RQ5:} What word embedding representatives in our model are close to each other in the semantic space?

\squishend
\subsection{Baselines and Our Models}
Since our methods are deterministic models, we compare them with state-of-the-art baselines in this direction.  
\squishlist
\item \textbf{SeqAttB:} Shang et al. \cite{shang2015neural} proposed a hybrid model that combines global scheme and local scheme \cite{bahdanau2014neural} to generate responses for original tweets on Sina Weibo. This model is one of the first work that generate responses for short text conversations.
\item \textbf{HRED:} It \cite{serban2015building} employs hierarchical RNNs for capturing information in a long context. HRED is a competitive method and a commonly used baseline for dialog generation systems.   
\item \textbf{our FCRG-BL:} This model uses the bilinear attention.
\item \textbf{our FCRG-DT:} This model uses the dot attention.
\squishend
\subsection{Experimental Settings}
\noindent\textbf{Data Processing.} Similar to \cite{serban2015building} in terms of text generation, we replaced numbers with $\texttt{<number>}$ and personal names with $\texttt{<person>}$. Words that appeared less than three times were replaced by $\texttt{<unk>}$ token to further mitigate the sparsity issue. 
Our vocabulary size was 15,321. The min, max and mean $|$tokens$|$ of the original tweets were 1, 89 and 19.1, respectively. The min, max and mean $|$tokens$|$ of FC-tweets were 3, 64 and 12.3, respectively. Only 791 ($1.2\%$) original tweets contained 1 token which is mostly a URL. 

\smallskip
\noindent\textbf{Experimental Design.} We randomly divided 73,203 pairs of the original tweets and FC-tweets into training/validation/test sets with a ratio of 80\%/10\%/10\%, respectively. The validation set was used to tune hyperparameters and for early stopping. At test time, we used the beam search to generate 15 responses per original tweet (beam size=15), and report the average results. To select the best hyperparameters, we conducted the standard grid search to choose the best value of a hidden size $H \in \{200, 300, 400\}$, and an output size $O\in \{256, 512\}$. We set word embedding size $D$ to 300 by default unless explicitly stated. The length of the original tweets and FC-tweets were set to the maximum value $N=89$ and $M=64$, respectively. The dropout rate was 0.2. We used Adam optimizer with fixed learning rate $\lambda=0.001$, batch size $b=32$, and gradient clipping was 0.25 to avoid exploded gradient. The same settings are applied to all models for the fair comparison.

A well known problem of the RNN-based decoder is that it tends to generate short responses. In our domain, examples of commonly generated responses were \textit{fake news url.}, \textit{you lie url.}, and \textit{wrong url.} Because a very short response may be less interesting and has less power to be shared (as we learned in Section \ref{sec:data_analysis}), we forced the beam search to generate responses with at least $\tau$ tokens. Since 92.4\% of FC-tweets had $|$tokens$|\geq5$, and 60\% FC-tweets had $|$tokens$|\geq10$, we chose $\tau \in \{0, 5, 10\}$. Moreover, as shown in Figure \ref{fig:length_vs_shares}, FC-tweets with $|$tokens$|$ $\in[10;20]$ usually had more shares than FC-tweets with $|$tokens$|$ $< 10$. In practice, fact-checkers can choose their preferred $|$tokens$|$ of generated responses by varying $\tau$. 

\smallskip
\noindent\textbf{Evaluation Metrics.} To measure performance of our models and baselines, we adopted several syntactic and semantic evaluation metrics used in the prior works. In particular, we used word overlap-based metrics such as BLEU scores \cite{papineni2002bleu}, ROUGE-L \cite{lin2004rouge}, and METEOR \cite{banerjee2005meteor}. These metrics evaluate the amount of overlapping words between a generated response and a ground-truth FC-tweet. The higher score indicates that the generated response are close/similar to the ground-truth FC-tweet syntactically. In other words, the generated response and the FC-tweet have a large number of overlapping words. Additionally, we also used embedding metrics (i.e. Greedy Matching and Vector Extrema) \cite{liu2016not}. These metrics usually estimate sentence-level vectors by using some heuristic to combine the individual word vectors in the sentence. The sentence-level vectors between a generated response and the ground-truth FC-tweet are compared by a measure such as cosine similarity. The higher value means the response and the FC-tweet are semantically similar.


\subsection{RQ1 \& RQ3: Quantitative Results based on Word Overlap-based Metrics}
In this experiment, we quantitatively measure performances of all models by using BLEU, ROUGE-L, and METEOR. Table \ref{tbl:performance_of_models_with_word_overlap_metrics} shows results in the test set. Firstly, our FCRG-DT and FCRG-BL performed equally well, and outperformed the baselines -- SeqAttB and HRED. In practice, FCRG-DT model is more preferable due to fewer parameters compared with FCRG-BL. Overall, our models outperformed SeqAttB perhaps because fusing global scheme (i.e. the last hidden state of the encoder) and output hidden state of every time-step $i^{th}$ in the encoder may be less effective than using only the latter one to compute context vector $\textbf{c}_j$. HRED model utilized only global context without using context vector $\textbf{c}_j$ in generating responses, leading to suboptimal results compared with our models.

Under no constraints on $|$tokens$|$ of generated responses, our FCRG-DT achieved 6.24\% ($p<0.001$) improvement against SeqAttB on BLEU-3 according to Wilcoxon one-sided test. In BLEU-4, FCRG-DT improved SeqAttB by 7.32\% and HRED by 7.76\% ($p<0.001$). In ROUGE-L, FCRG-DT improved SeqAttB and HRED by 3.32\% and 4.31\% with $p<0.001$, respectively. In METEOR, our FCRG-DT and FCRG-BL achieved comparable performance with the baselines.

When $|$tokens$|$ $\geq 5$, we even achieve better results. The improvements of FCRG-DT over SeqAttB were 7.05\% BLEU-3, 7.37\% BLEU-4 and 3.25\% ROUGE-L ($p<0.001$). In comparison with HRED, the improvements of FCRG-DT were 5.25\% BLEU-3, 5.64\% BLEU-4, and 2.97\% ROUGE-L $(p<0.001)$. Again, FCRG-DT are comparable with SeqAttB and HRED in METEOR measurement.

When $|$tokens$|\geq$ 10, there was a decreasing trend across metrics as shown in Table \ref{tbl:performance_of_models_with_word_overlap_metrics}. It makes sense because generating longer response similar with a ground-truth FC-tweet is much harder problem. Therefore, in reality, the Android messaging service recommends a very short reply (e.g., okay, yes, I am indeed) to reduce inaccurate risk. Despite the decreasing trend, our FCRG-DT and FCRG-BL improved the baselines by a larger margin. In particular, in BLEU-3, FCRG-DT outperformed SeqAttB and HRED by 17.9\% and 16.0\% $(p<0.001)$, respectively. For BLEU-4, the improvements of FCRG-DT over SeqAttB and HRED were 13.02\% and 11.74\% $(p<0.001)$, respectively. We observed consistent improvements over the baselines in ROUGE-L and METEOR.

Overall, our models outperformed the baselines in terms of all of the word overlap-based metrics. 

\begin{table}
	\caption{The results of human evaluation.}
	\vspace{-5pt}
	\begin{tabular}{llllc}
		\toprule[1pt]
		Opponent             & Win  & Loss & Tie  & Fleiss Kappa \\ \hline
		our FCRG-DT vs. SeqAttB & 40\% & 28\% & 32\% & 0.725        \\ \hline
		our FCRG-DT vs. HRED    & 40\% & 36\% & 24\% & 0.592        \\ \bottomrule[1pt]
	\end{tabular}
	\vspace{-5pt}
\label{tbl:human_evaluation}
\end{table}

\begin{table*}[t]
	\caption{Examples of original tweets, ground-truth FC tweets, generated responses of our FCRG-DT and two baselines.}
	\vspace{-5pt}
	\resizebox{1.0\linewidth}{!} {
		\begin{tabular}{l|l}
			\toprule[1pt]
			Pairs of the original tweet (\textbf{OT}) and ground-truth \centering \textbf{FC-tweet}                                                                                                                                                                                                                                      & Generated responses of our FCRG-DT and two baselines                                                                                                                                                                                                                                                                                \\ \hline
			\begin{tabular}[c]{@{}l@{}}\textbf{OT:} @user $<$person$>$ sold $<$number$>$ \% of our uranium \\ to russia . that's your candidate .  \\ \textbf{FC-tweet:} except that she did not . url @user\end{tabular}                                                                     & \begin{tabular}[c]{@{}l@{}}\textbf{SeqAttB}: @user no , $<$person$>$ did not start the birther movement . url \\ \textbf{HRED}:  @user here you go , read the whole thing . url \\ \textbf{FCRG-DT}:  @user allegations about hrc are false . \\     she did not have to give uranium to russia : url\end{tabular} \\ \hline
			\begin{tabular}[c]{@{}l@{}}\textbf{OT:} @user i just feel bad for poor franklin (?) , all alone on \\ his side of the table ... \\  \textbf{FC-tweet:} @user interesting . url\end{tabular}                                         & \begin{tabular}[c]{@{}l@{}}\textbf{SeqAttB}: @user there is no trucker strike in puerto rico url \\ \textbf{HRED}:  @user this is a fake . sorry folks . url \\ \textbf{FCRG-DT}: @user snopes says this is not true . url url\end{tabular}                                                                                                                            \\ \hline
			\begin{tabular}[c]{@{}l@{}}\textbf{OT:} @user that is why i asked the question so i could have \\ an understanding \\ \textbf{FC-tweet:} @user do use a troll as your source of info url\end{tabular}                                                                                       & \begin{tabular}[c]{@{}l@{}}\textbf{SeqAttB}:@user $<$person$>$ kissed by former klan member : url\\ \textbf{HRED}:  @user here you go , read the whole thing . url \\ \textbf{FCRG-DT}: @user this is a fake quote . sorry folks . url\end{tabular}                                                         \\ \hline

				\begin{tabular}[c]{@{}l@{}}\textbf{OT:} show me where $<$person$>$ actually said that , other \\ than this photoshopped picture from the internet . \\ \textbf{FC-tweet:} @user fake quote . url \end{tabular}                                                                       & \begin{tabular}[c]{@{}l@{}}\textbf{SeqAttB}: @user no, $<$person$>$ did not start the birther movement. url\\ \textbf{HRED}: @user yeah , no , they are not . url\\ \textbf{FCRG-DT}: @user this is a fake news site , right ? url\end{tabular}  \\ \hline
			
			\begin{tabular}[c]{@{}l@{}}\textbf{OT:} @user also , you are giving me a different set of statistics \\ than the ones i asked for . \\ \textbf{FC-tweet:} @user here's what he misquoted to start . url\end{tabular}                                                                       & \begin{tabular}[c]{@{}l@{}}\textbf{SeqAttB}: @user here you go , read the facts : url\\ \textbf{HRED}:  @user here ya go : url read this . url\\ \textbf{FCRG-DT}: @user this is a fake quote from $<$person$>$ . url\end{tabular}                                                                           \\ 
			  \bottomrule[1pt]
		\end{tabular}
	}
	\label{tbl:examples_generated_D_tweets}
	\vspace{-5pt}
\end{table*}

\subsection{RQ2 \& RQ3: Quantitative Results based on Embedding Metrics}
We adopted two embedding metrics to measure semantic similarity between generated responses and ground-truth FC-tweets \cite{liu2016not}.
Again, we tested all the models under three settings as shown in Table \ref{tbl:performance_of_models_with_word_overlap_metrics}. Our FCRG-DT performed best in all embedding metrics. Specifically, FCRG-DT outperformed SeqAttB by 3.98\% and HRED by 6.00\% improvements with $p<0.001$ in Greedy Matching. FCRG-DT's improvements over SeqAttB and HRED were 26.24\% and 5.62\% $(p<0.001)$, respectively in Vector Extrema. When $|$tokens$|\geq 5$, our FCRG-DT also outperformed the baselines in both Greedy Matching and Vector Extrema. 
In $|$tokens$|\geq10$, our models achieved better performance than the baselines in all the embedding metrics. In particular, FCRG-BL model performed best, and then FCRG-DT model was the runner up. To sum up, FCRG-DT and FCRG-BL outperformed the baselines in Embedding metrics.

\subsection{RQ4: Qualitative Evaluation}
Next, we conducted another experiment to compare our FCRG-DT with baselines qualitatively. In the experiment, we chose FCRG-DT instead of FCRG-BL since it does not require any additional parameters and had comparable performance with FCRG-BL. We also used $\tau=10$ to generate responses with at least 10 tokens in all models since lengthy responses are more interesting and informative despite a harder problem.

\smallskip
\noindent\textbf{Human Evaluation.} Similar to \cite{shang2015neural}, we randomly selected 50 original tweets from the test set. Given each of the original tweets, each of FCRG-DT, SeqAttB and HRED generated 15 responses. Then, one response with the highest probability per model was selected. We chose a pairwise comparison instead of listwise comparison to make easy for human evaluators to decide which one is better. Therefore, we created 100 triplets (original tweet, response$_1$, response$_2$) where one response was generated from our FCRG-DT and the other one was from a baseline. We employed three crowd-evaluators to evaluate each triplet where each response's model name was hidden to the evaluators. Given each triplet, the evaluators independently chose one of the following options: (i) win (response$_1$ is better), (ii) loss (response$_2$ is better), and (iii) tie (equally good or bad). Before labeling, they were trained with a few examples to comprehend the following criteria: (1) the response should fact-check information in the original tweet, (2) it should be human-readable and be free of any fluency or grammatical errors, (3) the response may depend on a specific case or may be general but do not contradict the first two criteria. The majority voting approach was employed to judge which response is better. If annotators rated a triplet with three different answers, we viewed the triplet as a tie. Table \ref{tbl:human_evaluation} shows human evaluation results. The Kappa values show moderate agreement among the evaluators. We conclude that FCRG-DT outperforms SeqAttB and HRED qualitatively.


\smallskip
\noindent\textbf{Case Studies.} Table \ref{tbl:examples_generated_D_tweets} presents examples of original tweets, ground-truth FC-tweets, and generated responses of the three models. Our FCRG-DT generated more relevant responses with clear fact-checking intention. For example, in the first example, FCRG-DT captured the \textit{uranium} in the original tweet and generated a relevant response. We observed that SeqAttB usually generated non-relevant content. Responses generated by FCRG-DT were more formal than ones generated by the baselines.  


\begin{figure}[t]
	\centering
	\subfigure[obamacare]{
		\label{fig:obamacare}
		\includegraphics[trim=40 40 25 40,width=0.3\linewidth,height=1in]{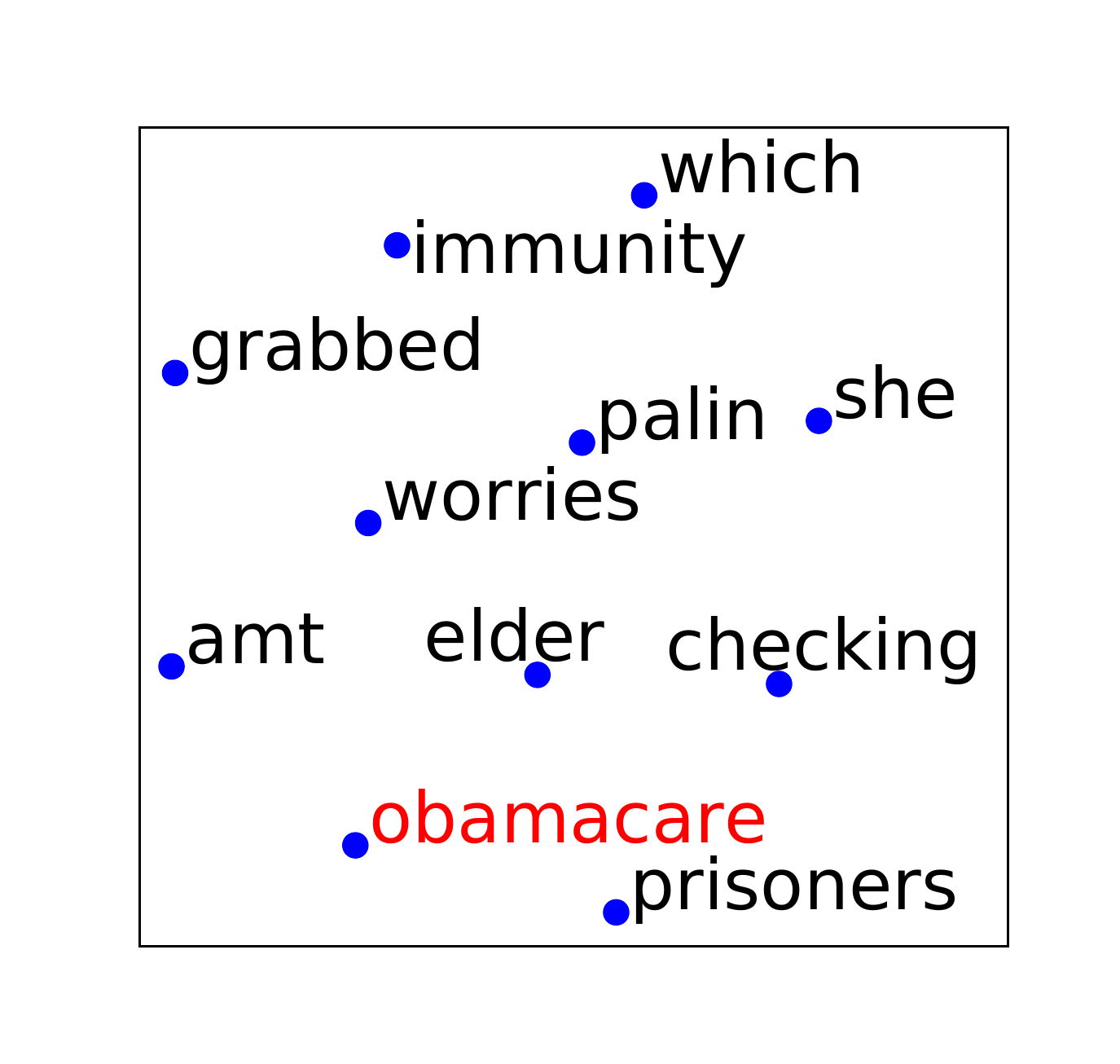}
	}
	\subfigure[politifact]{
		\label{fig:politifact}
		\includegraphics[trim=40 40 25 40,width=0.3\linewidth,height=1in]{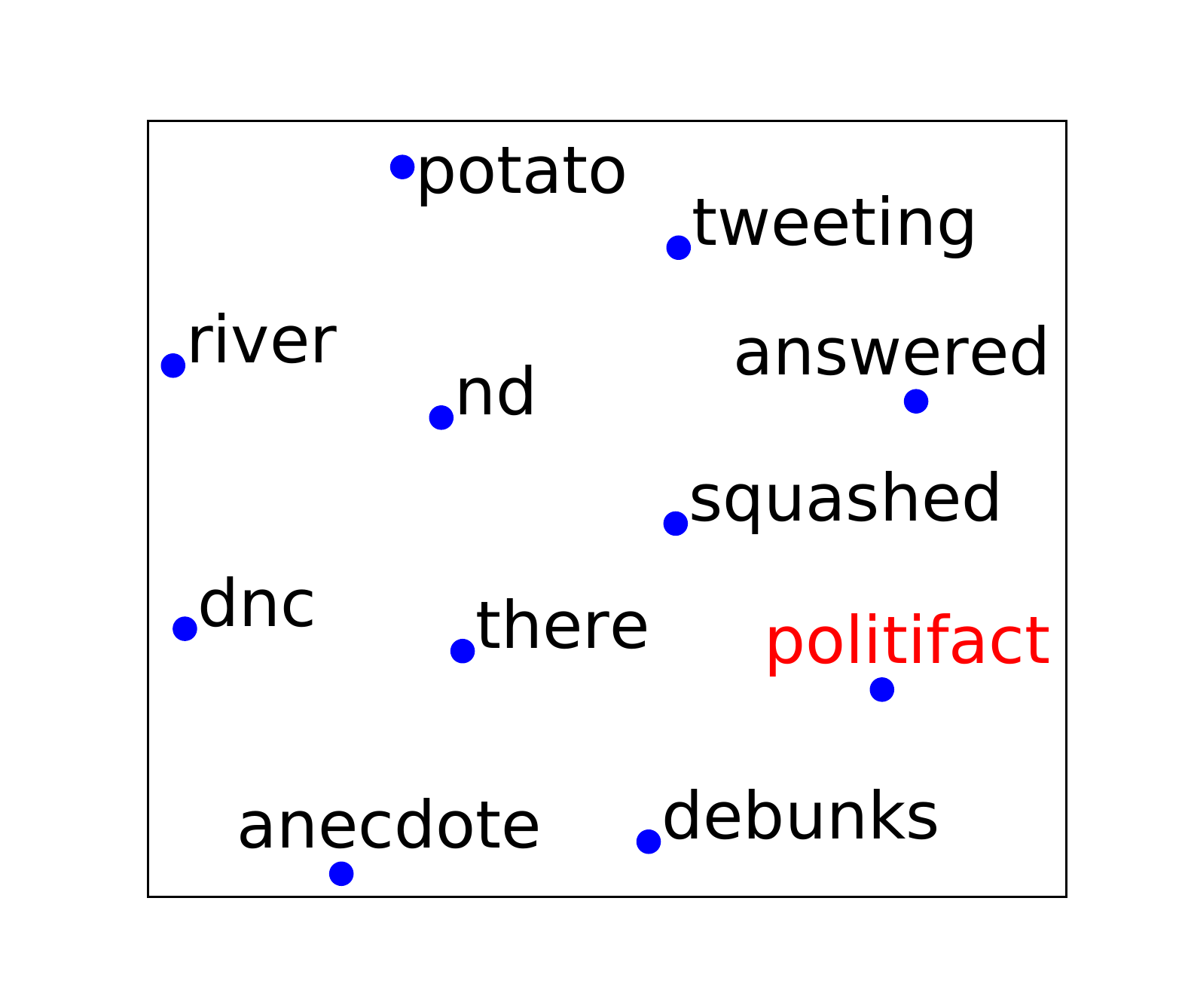}
	}
	\subfigure[anti-trump]{
		\label{fig:anti_trump}
		\includegraphics[trim=40 40 20 40,width=0.3\linewidth,height=1in]{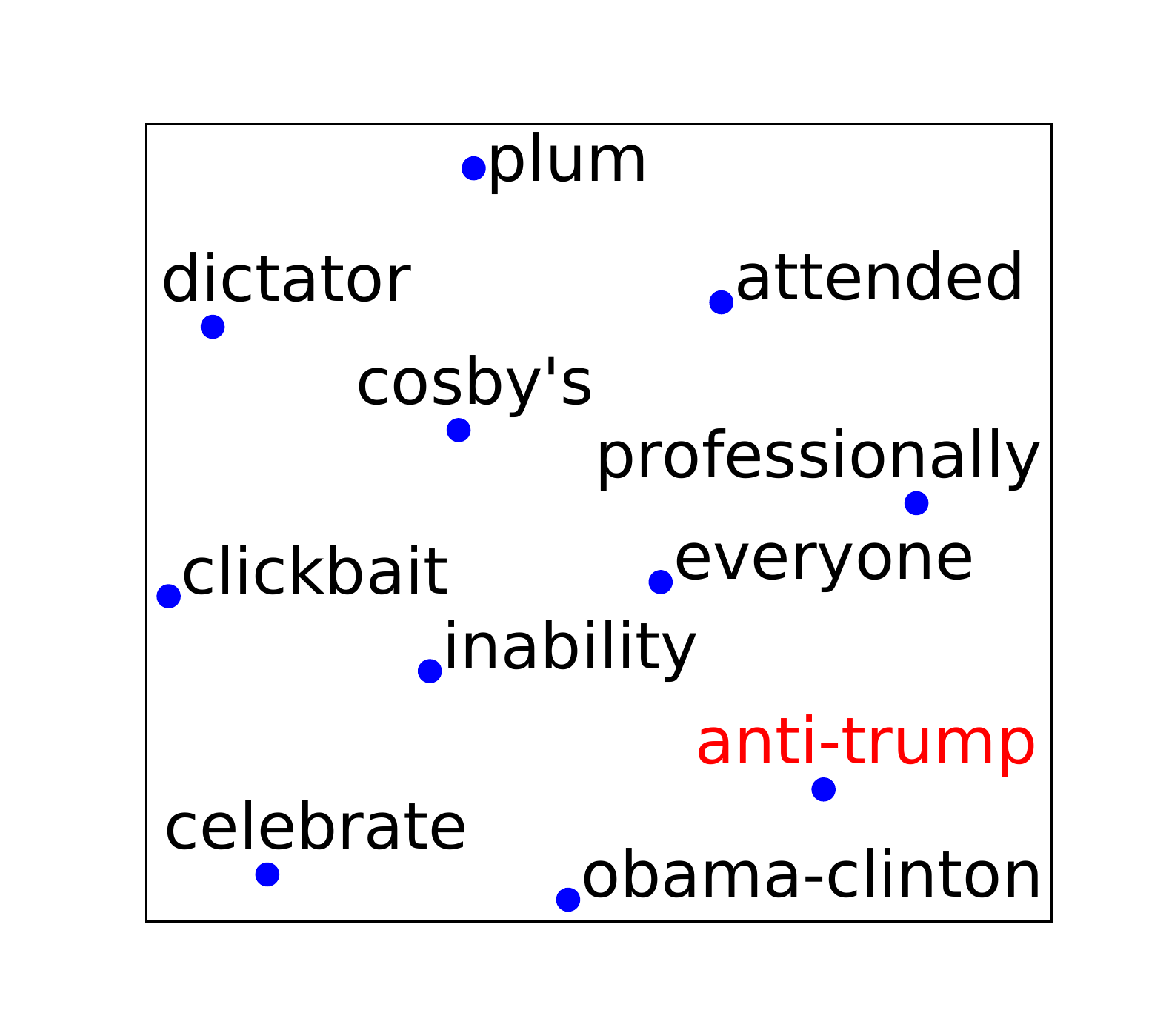}
	}
	\vspace{-5pt}
	\caption{2-D visualization of word embeddings learned from our FCRG-DT model.}
	\vspace{-5pt}
	\label{fig:word_embedding_viz}
	\vspace{-5pt}
\end{figure}

\subsection{RQ5: Similar Words in the Semantic Space}
As our word embeddings were trained from scratch, we seek to investigate if our model can identify words semantically close/similar to each other. This analysis will help us gain more insights about our dataset. After training our FCRG-DT, we extracted the embedding vectors from the shared embedding layer shown in Figure \ref{fig:framework_big_picture}. We selected three keywords such as \textit{obamacare}, \textit{politifact} and \textit{anti-trump}. For each keyword, we found top 10 most similar words based on cosine similarity between extracted embedding vectors and used t-SNE to project these vectors into 2D space as shown in Figure \ref{fig:word_embedding_viz}. Firstly, the keyword \textit{obamacare} associated with health care policies of the Obama administration in Figure \ref{fig:obamacare} was similar with \textit{elder}, \textit{AMT} (Alternative Minimum Tax), \textit{immunity} and \textit{checking}. Next, \textit{politifact} in Fig. \ref{fig:politifact} is close to \textit{debunks}, \textit{there}, \textit{anecdote}, \emph{DNC} (Democratic National Committee) and \textit{answered}. Finally, with \textit{anti-trump} in Fig. \ref{fig:anti_trump}, our model identified \textit{obama-clinton}, \textit{dictator}, \textit{clickbait}, and \textit{inability}. Based on this analysis, we conclude that embedding vectors learned from our model can capture similar words in the semantic space. 

\section{Discussion}
\label{sec:discussion}
Although our proposed models successfully generated responses with fact-checking intention, and performed better than the baselines, there are a few limitations in our work. Firstly, we assumed fact-checkers are freely choose articles that they prefer, and then insert corresponding fact-checking URLs into our generated responses. It means we achieved partial automation in a whole fact-checking process. In our future work, we are interested in even automating the process of selecting an fact-checking article based on content of original tweets in order to fully support fact-checkers and automate the whole process. Secondly, our framework is based on word-based RNNs, leading to a common issue: rare words are less likely to be generated. A feasible solution is using character-level RNNs \cite{kim2016character} so that we do not need to replace rare words with $\texttt{<unk>}$ token. In the future work, we will investigate if character-based RNN models work well on our dataset. Thirdly, we only used pairs of a original tweet and a FC-tweet without utilizing other data sources such as previous messages in online dialogues. As we showed in Figure \ref{fig:cosine_sim_boxplot}, FC-tweets often did not contain content of fact-checking articles, leading to difficulties in using this data source. We tried to use the content of fact-checking articles, but did not improve performance of our models. We plan to explore other ways to utilize the data sources in the future. Finally, there are many original tweets containing URLs pointing to fake news sources (e.g. breitbart.com) but we did not consider them when generating responses. We leave this for future exploration.  

\section{Conclusions}
In this paper, we introduced a novel application of text generation in combating fake news. We found that there were distinguishing linguistic features of FC-tweets compared with Normal and Random Replies. Notably, fact-checkers tended to refute information in original tweets and referred to evidences to support their factual corrections. Their FC-tweets were usually more formal, and contained less swear words and Internet slangs. These findings indicate that fact-checkers sought to persuade original posters in order to stop spreading fake news by using persuasive and formal language. In addition, we observed that when FC-tweets were posted, 7\% original tweets were removed, deleted or hidden from the public. After analyzing FC-tweets, we built a deep learning framework to generate responses with fact-checking intention. Our model FCRG-DT was able to generate responses with fact-checking intention and outperformed the baselines quantitatively and qualitatively. Our work has opened a new research direction to combat fake news by supporting fact-checkers in online social systems. 

\section*{Acknowledgment}
This work was supported in part by NSF grant CNS-1755536, AWS Cloud Credits for Research, and Google Cloud. Any opinions, findings and conclusions or recommendations expressed in this material are the author(s) and do not necessarily reflect those of the sponsors.

\appendix



%
%

\bibliographystyle{ACM-Reference-Format}
\bibliography{www}

\end{document}